%% file: 0_main.tex
\documentclass{article}

\usepackage{microtype}
\usepackage{graphicx}
\usepackage{subcaption}
\usepackage{booktabs} % for professional tables

\usepackage{hyperref}

\usepackage[accepted]{icml2026}

\usepackage{float}
\usepackage{booktabs}
\usepackage{graphicx}
\usepackage{caption}
\usepackage{tikz}
\usepackage{adjustbox}
\usepackage{enumitem}
\usepackage{comment}
\usepackage{textcomp}
\usepackage{amssymb}
\usepackage{mathtools}
\usepackage{amsmath}
\usepackage{amsthm}
\usepackage{xcolor}
\usepackage{subcaption} 
\usepackage{makecell}
\usepackage{wrapfig}
\definecolor{violet}{HTML}{956CB4}
\definecolor{green}{HTML}{6ACC64}
\definecolor{orange}{HTML}{EE854A}
\definecolor{red}{HTML}{D65F5F}
\usepackage{tabularx}
\usepackage{soul}

\newcommand{\blue}[1]{\textcolor{blue}{#1}}

\newcommand{\orange}[1]{\textcolor{orange}{#1}}

\usepackage[framemethod=TikZ]{mdframed}
\definecolor{lightGray}{gray}{0.97}
\definecolor{midGray}{gray}{0.65}
\definecolor{lightYellow}{RGB}{254,254,239}
\definecolor{darkGray}{rgb}{0.45,0.45,0.45}
\definecolor{darkerGray}{rgb}{0.3,0.3,0.3}
\definecolor{purpleBlue}{RGB}{249,250,254}
\definecolor{borderColor}{RGB}{199,201,208}
\mdfdefinestyle{myFrameStyle}{%
  linecolor=borderColor,
  linewidth=0.5pt,
  roundcorner=3pt,
  skipabove=5pt,
  skipbelow=0pt,
  innertopmargin=5pt,
  innerbottommargin=5pt,
  innerrightmargin=5pt,
  innerleftmargin=5pt,
  leftmargin=0pt,
  rightmargin=0pt,
  backgroundcolor=purpleBlue
}

\newcommand{\myFrame}[1]{
    \vspace{4pt}
    \begin{mdframed}[style=myFrameStyle,userdefinedwidth=
    \linewidth,align=center,skipabove=6pt,skipbelow=0pt]
    {#1}
    \end{mdframed}\vspace{-8pt}
}

\newcommand{\reb}[1]{%
    \textcolor{black}{#1}%
}

\usepackage[hang,flushmargin]{footmisc} % No indentation of footnotes, completely unnecessarily and eats up space

\usepackage[textsize=tiny]{todonotes}

\usepackage[textsize=tiny]{todonotes}

\usepackage[capitalize,noabbrev]{cleveref}

\theoremstyle{plain}

\theoremstyle{definition}

\theoremstyle{remark}

\usepackage[textsize=tiny]{todonotes}

\icmltitlerunning{Procedural Pretraining: Warming Up Language Models with Abstract Data}

\begin{document}

\twocolumn[
  \icmltitle{Procedural Pretraining: Warming Up Language Models with Abstract Data}



  \icmlsetsymbol{equal}{*}

  \begin{icmlauthorlist}
    \icmlauthor{Liangze Jiang}{equal,yyy,comp}
    \icmlauthor{Zachary Shinnick}{equal,sch}
    \icmlauthor{Anton van den Hengel}{sch}
    \icmlauthor{Hemanth Saratchandran}{sch}
    \icmlauthor{Damien Teney}{comp}
  \end{icmlauthorlist}
  
  \icmlaffiliation{yyy}{EPFL}
  \icmlaffiliation{comp}{Idiap Research Institute}
  \icmlaffiliation{sch}{AIML, Adelaide University}

  \icmlcorrespondingauthor{LJ}{liangze.jiang@epfl.ch}
  \icmlcorrespondingauthor{ZS}{zachary.shinnick@adelaide.edu.au}

  \icmlkeywords{Machine Learning, ICML}

  \vskip 0.3in
]



\printAffiliationsAndNotice{\hspace*{0.7em}\icmlEqualContribution} %

\input{0_abstract}
\input{1_introduction}

\input{7_related_work}

\input{2_methods}

\input{3_synthetic_reasoning}

\input{4_1_semantic_understanding}

\input{4_2_scaling_up}

\input{4_3_partial_transfer_semantic}

\input{6_mixtures}

\input{8_conclusion}

% % Acknowledgements should only appear in the accepted version.
% \section*{Acknowledgements}

% \textbf{Do not} include acknowledgements in the initial version of the paper

% If a paper is accepted, the final camera-ready version can (and usually should)
% include acknowledgements.  Such acknowledgements should be placed at the end of
% the section, in an unnumbered section that does not count towards the paper
% page limit. Typically, this will include thanks to reviewers who gave useful
% comments, to colleagues who contributed to the ideas, and to funding agencies
% and corporate sponsors that provided financial support.

% \clearpage

\section*{Impact Statement}

This paper presents work whose goal is to advance the field of 
Machine Learning. There are many potential societal consequences 
of our work, none of which we feel must be specifically highlighted here.

% Authors are \textbf{required} to include a statement of the potential broader
% impact of their work, including its ethical aspects and future societal
% consequences. This statement should be in an unnumbered section at the end of
% the paper (co-located with Acknowledgements -- the two may appear in either
% order, but both must be before References), and does not count toward the paper
% page limit. In many cases, where the ethical impacts and expected societal
% implications are those that are well established when advancing the field of
% Machine Learning, substantial discussion is not required, and a simple
% statement such as the following will suffice:

% ``This paper presents work whose goal is to advance the field of Machine
% Learning. There are many potential societal consequences of our work, none
% which we feel must be specifically highlighted here.''

% The above statement can be used verbatim in such cases, but we encourage
% authors to think about whether there is content which does warrant further
% discussion, as this statement will be apparent if the paper is later flagged
% for ethics review.

% % In the unusual situation where you want a paper to appear in the
% % references without citing it in the main text, use \nocite

\bibliography{ICML/references}
\bibliographystyle{icml2026}

%%%%%%%%%%%%%%%%%%%%%%%%%%%%%%%%%%%%%%%%%%%%%%%%%%%%%%%%%%%%%%%%%%%%%%%%%%%%%%%
%%%%%%%%%%%%%%%%%%%%%%%%%%%%%%%%%%%%%%%%%%%%%%%%%%%%%%%%%%%%%%%%%%%%%%%%%%%%%%%
% APPENDIX
%%%%%%%%%%%%%%%%%%%%%%%%%%%%%%%%%%%%%%%%%%%%%%%%%%%%%%%%%%%%%%%%%%%%%%%%%%%%%%%
%%%%%%%%%%%%%%%%%%%%%%%%%%%%%%%%%%%%%%%%%%%%%%%%%%%%%%%%%%%%%%%%%%%%%%%%%%%%%%%
\newpage
\appendix
\onecolumn
\input{9_appendix}

\end{document}

%% file: 0_abstract.tex
%\vspace{-3pt}
\begin{abstract}
% Pretraining directly on web-scale corpora is the de facto paradigm for building language models.
% We study an alternative setting
% where the model is initially exposed to abstract structured data,
% as a means to ease the subsequent acquisition of rich semantic knowledge,
% much like humans learn simple logic and mathematics before higher reasoning.
% We specifically focus on \textit{procedural data}, generated
% by formal languages and other simple algorithms, as such abstract data.

Pretraining language models directly on web-scale corpora is the de facto paradigm.
We study an alternative
where the model is initially exposed to \textit{abstract structured data} to ease the subsequent acquisition of rich semantic knowledge,
much like humans learning simple logic and mathematics before higher reasoning.
We focus on \textit{procedural data}, generated
by formal languages and other simple algorithms, as such abstract data.

\vspace{-0.2em}
% \textbf{Method and findings.}
We first 
% use small models to 
diagnose the algorithmic skills that
different forms of procedural data can improve, often significantly.
For example, the accuracy of context recall (\textsc{Needle-in-a-haystack}) jumps from $10$ to $98\%$ when a model is pretrained on Dyck sequences (balanced brackets).
Second, we study how these gains are reflected in pretraining larger models (up to 1.3B).
% We find that procedural pretraining significantly improves performance on natural language, code, and informal mathematics
% (\textsc{C4}, \textsc{CodeParrot}, and \textsc{DeepMind-Math} datasets),
% by front-loading as little as $0.1$\% procedural data.
We find that front-loading as little as $0.1$--$0.3$\% procedural data significantly outperforms standard pretraining on natural language, code, and informal mathematics
(\textsc{C4}, \textsc{CodeParrot}, and \textsc{DeepMind-Math} datasets).
Notably, this also enables the models to reach the same loss value
with only $55$\,/\,$67$\,/\,$86$\% of the original data and thus a comparable reduction in FLOPs. %  of these datasets.
Third, we explore the mechanisms behind the benefits and find that procedural pretraining instils non-trivial structure in both attention and MLP layers. 
The former is particularly important for structured domains (e.g.\ code), and the latter for language.
%while the latter is crucial for language interpretation. % D: too long
Finally, we lay a path for combining multiple forms of procedural data. 
Our results show that procedural pretraining is a simple, lightweight means to improving performance and accelerating language model pretraining, ultimately suggesting the promise of disentangling knowledge acquisition from reasoning in LLMs.
Code is available at \href{https://zlshinnick.github.io/procedural-pretraining-page/}{this page}.
% \red{\textbf{Implications.} 
% %A We add evidence and explanations to the literature on procedural pretraining, showing benefits in more domains and settings.
% Procedural pretraining is a
% remarkably simple means of
% improving performance and
% speeding up training for language models.
% %reducing training data requirements of transformers. % D: A few words too long
% It 
% %The analysis of where its impact lies in transformers % D: too long
% ultimately suggests the possibility of disentangling the acquisition of knowledge from reasoning in LLMs.}
\end{abstract}

%% file: 1_introduction.tex
\vspace{1pt}
\section{Introduction}
\label{sec:introduction}
\vspace{2pt}

Large language models (LLMs) simultaneously acquire multiple forms of knowledge during pretraining.
They absorb rich semantic content, but also acquire abilities for
manipulating this knowledge.
%}, connecting related facts, producing coherent natural language, and even generating highly-structured outputs such as code.
This entangled learning of knowledge and abstract skills has been identified as a key limitation of current models \citep{han2025position,kumar2025questioning},
% such as their tendency to rely on
leading to their reliance on surface-level heuristics rather than
systematic reasoning procedures~\citep{nikankin2025arithmetic}.

% \textbf{Pretraining with procedural data.}
To mitigate knowledge-reasoning entanglement,
we study \textit{using abstract, structured
data to `warm up' language models}.
Intuitively, this is a lightweight pretraining that  
builds algorithmic scaffolding
% teach abstract elementary operations
without relying on semantic shortcuts, much like
infants learning games like stacking blocks~\citep{smith2005development}
before moving to sophisticated reasoning and knowledge.
With \textbf{procedural pretraining}, we posit that early exposure of LMs to procedural data\footnote{We use procedural data to refer to the output of
explicit algorithms (e.g.\ formal languages or sorting), in contrast to synthetic data, which is typically generated by trained models such as LLMs.}
facilitates and enhances standard pretraining on semantically-rich corpora.

In prior work, \citet{hu2025between} showed that ``pre-pretraining'' LLMs on data generated from formal languages yields more value per token than natural language.
\citet{wu2022insights} and \citet{zhang2024intelligence}
successfully used data from simple algorithms and cellular automata.
Their findings echo the established practice of pretraining on computer code, another structured domain thought to aid learning compositional and recursive reasoning~\citep{petty2024does}. 
Prior works, however, typically treat procedural data as either \textit{imitation} of linguistic properties, or as a drop-in \textit{substitute} of standard pretraining.
In contrast, we study procedural data from a broader \textit{algorithmic} view, and position it explicitly as a \emph{complementary} warming-up stage for standard pretraining. 
Our experiments contain two main parts.
The \textbf{first part} identifies why and when procedural pretraining helps using algorithmic tasks as diagnostic tools, while the \textbf{second part} shows, with larger models, procedural pretraining improves standard pretraining in several domains of practical interest across scales. Our contributions are:
%A\newpage

\begin{figure*}[t!]
    \centering
        \includegraphics[width=\linewidth]{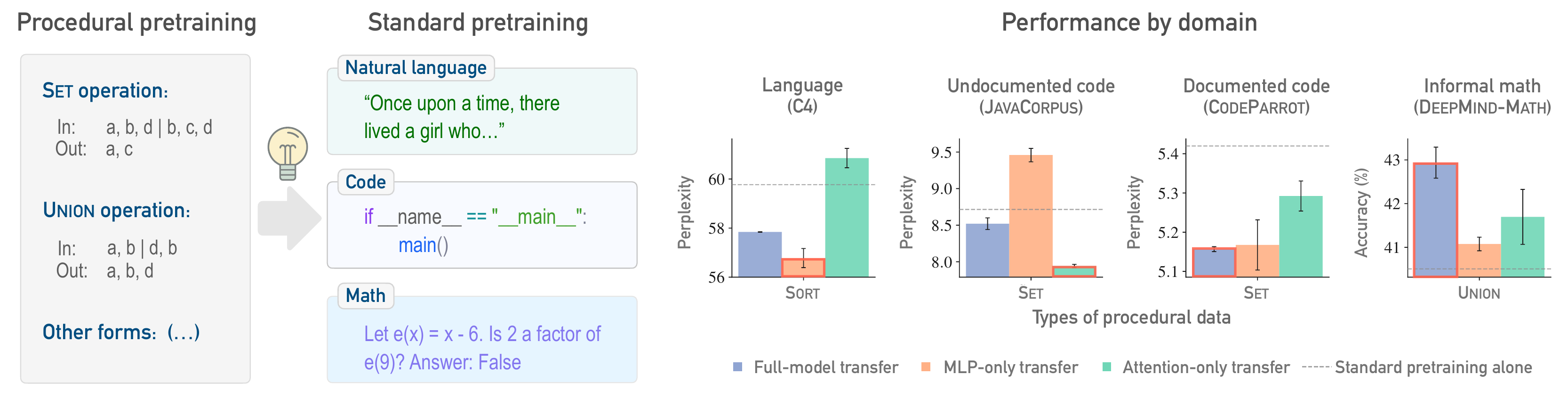}
        \caption{%
        \textbf{(Left)}~We pretrain language models on procedural data
        before exposing them to
        standard datasets of language, code, or mathematics.
        The procedural data is generated with simple algorithms 
        and aims to teach elementary skills to aid the
        acquisition of semantic knowledge.
        \textbf{(Right)}~This lightweight initial step
        speeds up standard pretraining
        and improves performance on diverse domains,
        % by transferring specific pretrained layers (MLP vs.\ attention).
        %\textbf{(Right)}~This pretraining only needs to be lightweight (not pictured) to %provide improvements across diverse domains,
        with different pretrained layers (MLP vs.\ attention) contributing differently to each domain.
        }
    \label{fig:teaser}
    \vspace{-3pt}
\end{figure*}

% \footnotetext{We use procedural data to refer to the output of
% explicit algorithms (e.g.\ formal languages), in contrast to synthetic data, which is typically generated by trained models such as LLMs.}

% \newpage
% \red{Our contributions push the use of procedural pretraining in four directions.}
% \vspace{-2pt}

\textbf{(1)~Probing procedural pretraining with algorithmic tasks.}
% We evaluate the effects of different forms of procedural data
% and find that they each enhance specific algorithmic skills
We find that different forms of procedural data each enhance specific algorithmic skills (Section~\ref{sec:algorithmic_reasoning1}).
%, e.g.\ arithmetic or context recall
%reasoning tasks, such as needle-in-a-haystack and decimal addition 
The pretrained information also proves to be %modular and
localised in specific layers (attention vs.\ MLPs, Section~\ref{sec:algorithmic_reasoning2}).
We also rule out simplistic explanations that could account for the observed improvements,
%, and reflect precise weight structure rather than artifacts
such as rescaling the initialisation or a generic attention sharpening
 (Section~\ref{sec:algorithmic_reasoning3}).
% We also demonstrate that procedural pretraining imparts precise weight structure by ruling out other explanations such as weight scaling or generic attention sharpening (Section~\ref{sec:algorithmic_reasoning3}).

\textbf{(2)~Transfer to pretraining on diverse domains.}
We show that the improvements on algorithmic skills transfer to
multiple semantic domains, namely natural language, code, and informal mathematics
(Section~\ref{sec:semantic_understanding1}--\ref{sec:scaling_up}).
The information learned in procedural pretraining proves to be
\textit{complementary} to standard pretraining datasets.
For example, we consistently improve over standard pretraining with as little as
$0.1$\,--\,$0.3$\% extra procedural tokens.
Procedural data also proves to be an \textit{efficient substitute} to standard data.
On \textsc{C4}, \textsc{CodeParrot}, and \textsc{DeepMind-Math} datasets,
it enables models to reach the same loss with respectively
$55$\%, $67$\%, and $86$\% of the original data and therefore a comparable FLOP reduction. 
\reb{Furthermore, we validate these findings across different model sizes (up to 1.3B parameters) and data sizes (up to 10.5B tokens), and show that the gains at standard pretraining persist on downstream language, code generation and commonsense reasoning tasks (Section~\ref{sec:downstream}).}

% \textbf{(3) Discovering layer specialisation} %in procedural pretraining}
\textbf{(3)~Localising transferable pretrained information}
(Section~\ref{sec:partial_transfer_semantic}).
We localise useful procedurally-pretrained information for each domain.
We find that the attention layers are more important for structured, language-free domains like pure code,
while MLP layers help natural language more.
The latter is intriguing because MLPs are believed to store factual knowledge in LLMs 
\citep{dong2025attention,geva2020transformer,xu2025filtering}
which procedural data cannot directly provide.
On data containing both natural language and structured data such as \textsc{CodeParrot} (\textit{documented} code) and \textsc{DeepMind-Mathematics} (\textit{informal} mathematics), both components are important
(Figure~\ref{fig:teaser} right). 
% \reb{These results reveal further performance gain by selectively transferring procedurally-pretrained weights for the subsequent training.}

\textbf{(4) Combining the benefits of different forms of procedural data} (Section~\ref{sec:mixture}).
We explore two techniques
and obtain promising results by either
pretraining on a mixture of data types,
or surgically combining weights of several pretrained models.
This lays out several directions for future work.

% \red{These results show that procedural data is
% a data-efficient alternative for pretraining, 
% and a complementary source of knowledge to standard datasets.
% We discuss in Section~\ref{sec:discussion} how this line of work
% may ultimately help disentangle the acquisition of knowledge from reasoning in LLMs.}

Our results show that procedural data is both
a data-efficient alternative and an effective complement to standard pretraining.
We discuss intriguing future directions in Section~\ref{sec:discussion} and how our findings
may ultimately improve the knowledge and reasoning acquisition in LLMs.

%% file: 7_related_work.tex
\section{Related Work}
\label{sec:related_work}

The linguistic literature contains a number of results
on training language models with artificial data.
These works often use formal languages to imitate properties of natural language
\citep{chiang2022transferability,goodale2025meta,mccoy2023modeling,papadimitriou2023injecting,ri2022pretraining,hu2025between}.
\reb{In contrast, we follow a more general algorithmic perspective, and find how different types of procedural data can improve specific algorithmic skills.}
Hence, we also study benefits on domains beyond language, namely code and informal mathematics.

We use procedural data generated with simple algorithms and cellular automata, as in recent works
\citep{lindemann2024sip,wu2022insights,wu2021lime,zhang2024intelligence,bloem2025universalpretrainingiteratedrandom}. 
These works focus on procedural data as a \emph{substitute} for standard pretraining data.
In contrast, we also evaluate procedural data as a \emph{complement},
and find that it can impart capabilities lacking from standard semantic data \reb{across diverse domains}. 
\reb{Additionally, we validate empirically that the benefits of procedural pretraining even persist after further fine-tuning on downstream tasks.} %, which is an important practical result that was absent from prior work.}
We also analyse in greater depth the mechanisms behind the empirical benefits,
such as the localisation of pretrained knowledge in MLP vs.\ attention layers.
\reb{This sometimes also reveals further empirical pretraining gains by only transferring specific layers from procedural pretraining.}
% (see Section~\ref{sec:experimental_setup} for definition).
Finally, while most of this existing work focuses on a single type of data, we take steps towards combining multiple types of procedural data, \reb{which lays out a path for important next steps for this line of work}.

A concurrent work~\citep{shinnick2025can} shows that procedural data benefits visual learning. 
Together with our findings, this implies that procedural data might inject modality-agnostic mechanisms~\citep{huh2024platonic} into the model. % This concurs with our experiments in ruling out simplistic explanations of procedural data's benefits.
We provide an extended discussion on other related work in Appendix~\ref{app:related_work}.

%% file: 2_methods.tex
\vspace{-3pt}
\section{Preliminaries}
\label{sec:methodology}
\vspace{-2pt}

We use the following terminology throughout this paper.
%We summarize the terminology and then describe the experimental and procedural data setup.
\begin{itemize}[leftmargin=*, itemsep=2pt, parsep=0pt, topsep=-3pt]
\item \textbf{Procedural pretraining}: the initial ``pre-pretraining'' of a model to procedural data,
before other stages such as \textbf{standard pretraining} with semantic data.
%using natural language, code, mathematics, etc.
\item \textbf{Procedural data}: non-semantic data generated from a simple algorithm, for example
formal languages, cellular automata, or other rule-based processes in
%a Set (deduplication) algorithm, or the simulation of a stack memory 
Section~\ref{subsec:methodology-procedural}.
% This differs from \textbf{synthetic data}, which refers in the current literature to data generated by learned models such as another LLM. % L: seems not needed since it's quite clear at this point.
\item \textbf{Algorithmic data}: data/tasks for diagnostic purposes to probe basic capabilities, such as needle-in-a-haystack, addition, etc. described in Section~\ref{sec:algorithmic_reasoning}.
\item \textbf{Semantic data}: 
standard (semantic) data used to pretrain language models, for example natural language, computer code, or informal mathematics.
\end{itemize}

\vspace{-2pt}
\subsection{Experimental Setup}
\label{sec:experimental_setup}

% \todo{D: Remove this paragraph to save space??? L: yes we could remove given it's somehow mentioned in the caption of figure 1 and elsewhere. Let's see how tight the space is.}
% We investigate the benefits of procedural pretraining.
% Intuitively, this implements
% the same idea as the common paradigm in machine learning
% of pretraining then fine-tuning
% i.e.\ progressing from general to specific knowledge.
% With procedural pretraining, we initial expose the model to abstract data,
% aiming to teach general algorithmic priors
% before exposure to standard semantic data.

We train \textsc{GPT-2}-type decoder-only transformers from scratch
with a standard next-token prediction objective \citep{radford2019language} (see Appendices~\ref{app:model_details}\,and\,\ref{app:experimental-details} for details.). 
When pretraining on procedural data that involves input/output pairs (Section~\ref{subsec:methodology-procedural}), we compute the loss only on output tokens.
Apart from Section~\ref{sec:mixture}, each experiment uses a single type of procedural data.

\reb{\textbf{Data setup.}} We first train each model on $T_1$ procedural tokens,
then on $T_2$ standard tokens from the target data.
The target data is either an algorithmic task in Section~\ref{sec:algorithmic_reasoning} for diagnostic purpose,
or a semantic dataset in Section~\ref{sec:semantic_understanding} reflecting standard training practices.
The \reb{\textbf{baseline}} is the same model trained with no procedural data ($T_1=0$).
We adjust $T_1$ and $T_2$ following either of these two settings:
%Specifically, we show two settings \red{(as shown in Figure~\ref{fig:teaser})}:
\begin{itemize}[leftmargin=*, itemsep=3pt, parsep=0pt, topsep=-2pt]
    \item \textbf{Additive setting.} We keep $T_2$ fixed and vary $T_1$ to measure the performance gain of \emph{additional} procedural tokens. This evaluates whether procedural data provides a training signal that semantic data alone does not impart.
    %inductive biases or structural knowledge that semantic data alone does not efficiently impart.
    \item \textbf{Substitutive setting.} We reduce $T_2$ while increasing
    $T_1$ (by a much smaller amount) to match the baseline performance.
    This evaluates how procedural pretraining can be a cheaper substitute for standard pretraining.
% \todo{Z: Do we need "typically", since it is always a much smaller amount}

\end{itemize}\vspace{-1pt}
\reb{\textbf{Weight setup.}} All the weights of the model are trained in \reb{both procedural pretraining and any subsequent training stages, i.e.\ nothing is frozen}.
Each experiment uses either of the two following transfer settings 
between the two phases.
\begin{itemize}[leftmargin=*, itemsep=3pt, parsep=0pt, topsep=-2pt]
    \item \textbf{Full-model transfer.} The standard practice, i.e.\ using all procedurally-pretrained weights.\footnotemark
    \item \textbf{Selective attention-only or MLP-only transfer.}
    We only use the pretrained weights of selected layers and reinitialize
    others to random values.
    This evaluates where useful pretrained information is stored,
    motivated by the evidence that MLP and attention layers perform different computations
    \citep{dong2025attention,xu2025filtering}.
\end{itemize}\vspace{-1pt}

\footnotetext{In Sections~\ref{sec:semantic_understanding} and \ref{sec:mixture}
 we reinitialise the token embeddings to random values since there is no correspondence between the vocabularies of procedural and semantic data. In
Section~\ref{sec:algorithmic_reasoning}
(procedural $\!\rightarrow\!$ algorithmic transfer), we instead initialise embeddings to the mean vector, as there is no semantic domain shift.}

\subsection{Generating Procedural Data}
\label{subsec:methodology-procedural}

Each procedural data type is defined by a data-generating algorithm.
%(see Appendix~\ref{app:procedural-pretraining} for details).
We use algorithms that produce structurally rich data
where next-token prediction requires precise symbol manipulation, compositional reasoning, and/or long-range dependency tracking.
We select these from prior work
and also introduce a new one (\textsc{Stack}). 
See Figure~\ref{fig:procedural_data_examples} for examples.
They each takes hyperparameters detailed in Appendix~\ref{app:procedural-pretraining} and generates \textit{short sequences} (max. $128$ tokens).
%such as the generated sequence length.
%They are non-deterministic and thus allow sampling data virtually infinitely without repetition.

\begin{itemize}[leftmargin=*, itemsep=3pt, parsep=0pt, topsep=-5pt]
\item
\textbf{Sequence transformations.} A random sequence is presented and the model must predict its transformed version \citep{wu2022insights}.
%followed by a transformed version. The training loss is applied only on the transformed version.
%require the model to perform a specific transformation on an input sequence.
This includes
\textsc{Set} (token deduplication),
\textsc{Reverse} (reversing the input),
\textsc{Identity} (copying the input),
\textsc{Union} (ordered combination of two sequences with duplicates removed),
\textsc{Sort} (copy in ascending order)
and
\textsc{Delete} (removal of a specified token).

\item
\textbf{Memory operations.} \textsc{Stack} simulates a stack memory, tracking state over a random series of push and pop operations. The model must predict the final
memory contents from top to bottom.
%\todo{D: "value on top of the stack"?
%Z: output is the remaining elements of the stack from top to bottom.}

\item
\textbf{Formal languages.}
We use classical formal languages for balanced parentheses
\citep{hu2025between, papadimitriou2023injecting},
%We also explore the unique capabilities instilled by other established tasks from prior %work. These include the formal languages of
\textsc{k-Dyck} (nested) and \textsc{k-Dyck Shuffle} (non-nested).
The model is trained for next-token prediction to generate sequences from the target language, and we vary $k$ to control the complexity of the nesting. 
%The underlying data are drawn from the formal languages. The model is trained in a generative setting: given a sequence prefix, it must predict the next token, thereby producing strings from the target language.

\item
\textbf{Cellular automata.}
We use the elementary cellular automaton \textsc{Eca rule 110} following 
\citet{zhang2024intelligence}, where a binary sequence evolves via deterministic Markovian dynamics.
Each sequence describes a random state of the ECA and the model must predict the next state.
\end{itemize}\vspace{-1pt}

%% file: 3_synthetic_reasoning.tex
\section{Probing Procedural Pretraining with Algorithmic Reasoning}
\label{sec:algorithmic_reasoning}

% \begin{figure*}[t!]
%     \vspace{-3pt}
%     \centering
%     \setlength{\tabcolsep}{1pt}
%     \begin{tabularx}{\linewidth}{*{5}{>{\centering\arraybackslash}X}}
%         \scriptsize\phantom{mmm}\textsc{Haystack}&
%         \scriptsize\phantom{mmm}\textsc{Addition}&
%         \scriptsize\phantom{mmm}\textsc{Reversed add.}&
%         \scriptsize\phantom{mmm}\textsc{Multiplication}&
%         \scriptsize\phantom{mmm}\textsc{Sorting}
%     \end{tabularx}\vspace{1pt}
%     \includegraphics[trim=0pt 8pt 0pt 32.5pt,clip,width=1.0\linewidth]{figures/final/alg_reasoning_all_tasks_2.pdf}
%     % Trim bottom (white space)/top (ugly titles) of PDF file
%     \vspace{-12pt}
%     \caption{
%     \textbf{Different types of procedural pretraining can significantly improve over standard training} (dashed line) across various algorithmic tasks.
%     If we remove the structure within the procedural data by shuffling the sequences
%     (\textit{Best model shuffled}), the performance falls to the baseline.
%     Reported values are the means over 10 seeds (full results with variance in Appendix~\ref{app:more_results-algorthmic-reasoning}).
%     }
%     \label{fig:alg_reasoning_results}
% \end{figure*}

\begin{figure*}[t!]
    \vspace{-3pt}
    \centering
    \setlength{\tabcolsep}{1pt}

    \begin{tabularx}{\linewidth}{*{5}{>{\centering\arraybackslash}X}}

        \scriptsize
        \begin{tabular}{c}
            \textsc{Haystack} \\
            {\tiny \textsc{(Best: 16 Dyck)}}
        \end{tabular}
        &

        \scriptsize
        \begin{tabular}{c}
            \textsc{Addition} \\
            {\tiny \textsc{(Best: 16 Dyck)}}
        \end{tabular}
        &

        \scriptsize
        \begin{tabular}{c}
            \textsc{Reversed Add.} \\
            {\tiny \textsc{(Best: ECA)}}
        \end{tabular}
        &

        \scriptsize
        \begin{tabular}{c}
            \textsc{Multiplication} \\
            {\tiny \textsc{(Best: Union, Delete)}}
        \end{tabular}
        &

        \scriptsize
        \begin{tabular}{c}
            \textsc{Sorting} \\
            {\tiny \textsc{(Best: 16 Dyck, Set)}}
        \end{tabular}

    \end{tabularx}

    \vspace{-10pt}

    \includegraphics[
        trim=0pt 8pt 0pt 32.5pt,
        clip,
        width=1.0\linewidth
    ]{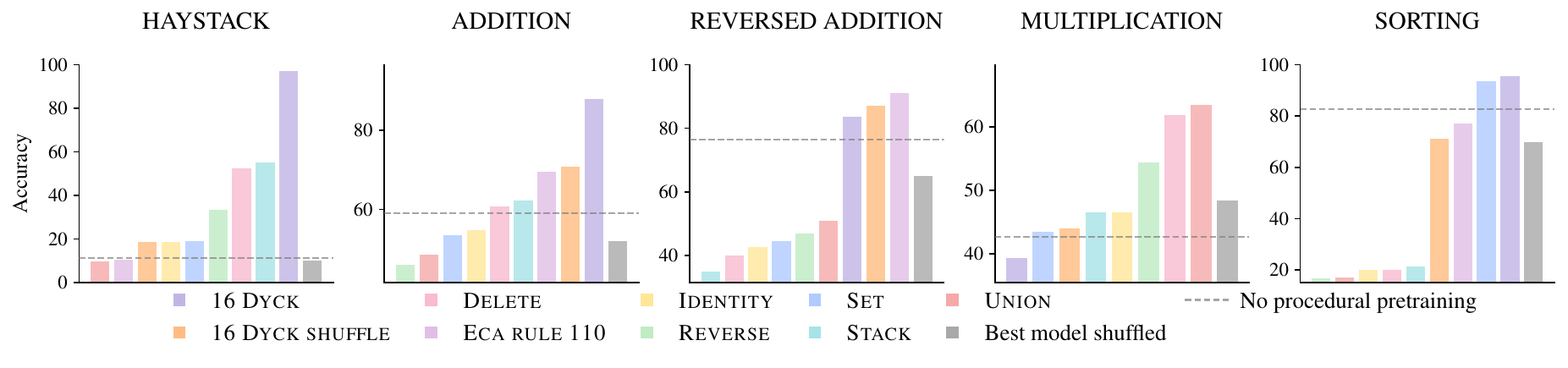}

    % Trim bottom (white space)/top (ugly titles) of PDF file
    % \vspace{-4pt}

    \caption{
    % \textbf{Different types of procedural pretraining can significantly improve over standard training} (dashed line) across various algorithmic tasks.
    \textbf{Different types of procedural pretraining facilitate learning different algorithmic skills.}
    If we remove the structure within the procedural data by shuffling the sequences
    (\textit{Best model shuffled}), the performance falls to the baseline. We sort the types of data by increasing performance on each task (\textit{Best model shuffled} always last, \textit{best performers} mentioned in the titles).
    Reported values are the means over 10 seeds (full results with variance in Appendix~\ref{app:more_results-algorthmic-reasoning}).
    }

    \label{fig:alg_reasoning_results}

\end{figure*}

We first train small transformers (two layers, four attention heads) on
specific types of procedural data,
then fine-tune them on common algorithmic tasks to evaluate how specific types of procedural data improve the following skills
(training and test data are i.i.d.; full details are in Appendix~\ref{app:algorithmic_task_details}).

\begin{itemize}[leftmargin=*, itemsep=3pt, parsep=0pt, topsep=-5pt]
\item
\textbf{Memory recall.} The needle-in-a-haystack task (\textsc{Haystack}) evaluates long-context retrieval. 
Each input has $30$ key–value pairs \(([m_1,c_1,\ldots,m_k,c_k])\) and a query marker $m_u$; 
the model must output the associated value $c_u$. %  associated with $m_u$. 
Accuracy is measured on the retrieved token. 

\item
\textbf{Arithmetic.} We evaluate three tasks.
\textsc{Addition} sums two $5$-digit integers (\texttt{a+b=}), requiring right-to-left carry propagation, opposite to the autoregressive order.
\textsc{Reversed addition} uses $10$-digit numbers with reversed inputs and outputs, aligning carries with autoregression.
\textsc{Multiplication} computes the product of two $5$-digit integers (\texttt{a$\times$b=}), predicting only result digits.
All tasks are tokenized per digit, and the accuracy is measured over the output digits.

\item
\textbf{Logical and relational processing.} With \textsc{Sorting}, the model receives $10$
integers from $[0,\!99]$ and a separator, and outputs the sorted sequence.  
The accuracy is computed on the output tokens.  

\end{itemize}\vspace{-1pt}
The model size is kept small here to avoid performance saturation through scale and to enable reliable conclusions by running each experiment with $10$ different seeds.
%-------------------------------------------------------------------%
\subsection{Which Algorithmic Skills Improve with Procedural Pretraining?}
\label{sec:algorithmic_reasoning1}

\textbf{Setup.} We use the \textit{additive} setting in Section~\ref{sec:experimental_setup}: for every combination of a type of procedural data
and algorithmic task, we train on $T_1$ procedural tokens then $T_2$ tokens of the algorithmic task. The baseline model uses $T_1\!=\!0$.

\textbf{Results.}
Figure~\ref{fig:alg_reasoning_results} shows that many types of procedural data
significantly improve performance on various tasks.
The best type of procedural data varies across task.
For example, pretraining on \textsc{k-Dyck} improves context recall (\textsc{Haystack}),
while \textsc{Eca rule 110} benefits \textsc{Reversed Addition}.
This indicates that each type of procedural data improves different skills.
We also evaluate the best model pretrained on \emph{randomly shuffled} procedural sequences.
This conserves the token distribution within sequences while disrupting their structure
(\textit{Best model shuffled} in Figure~\ref{fig:alg_reasoning_results}).
% The performance then falls below the baseline model.
The performance subsequently drops back to baseline.
This shows that
the structure in the procedural data is essential. 

% \begin{figure*}[t!]
%     \vspace{-3pt}
%     \centering
%     \setlength{\tabcolsep}{1pt}
%     \begin{tabularx}{\linewidth}{*{5}{>{\centering\arraybackslash}X}}
%         \scriptsize\phantom{mmm}\textsc{Haystack}&
%         \scriptsize\phantom{mmm}\textsc{Addition}&
%         \scriptsize\phantom{mmm}\textsc{Reversed add.}&
%         \scriptsize\phantom{mmm}\textsc{Multiplication}&
%         \scriptsize\phantom{mmm}\textsc{Sorting}
%     \end{tabularx}\vspace{1pt}
%     \includegraphics[trim=0pt 8pt 0pt 32.5pt,clip,width=1.0\linewidth]{figures/final/alg_reasoning_all_tasks_2.pdf}
%     % Trim bottom (white space)/top (ugly titles) of PDF file
%     \vspace{-15pt}
%     \caption{
%     \textbf{Different types of procedural pretraining can significantly improve over standard training} (dashed line) across various algorithmic tasks.
%     If we remove the structure within the procedural data by shuffling the sequences
%     (\textit{Best model shuffled}), the performance falls to the baseline.
%     Reported values are the means over 10 seeds (full results with variance in Appendix~\ref{app:more_results-algorthmic-reasoning}).
%     }
%     \label{fig:alg_reasoning_results}
% \end{figure*}

\myFrame{\textbf{Take-away.} Among different types of procedural data, each improves specific algorithmic skills.}

% \damien{Removed: "which we hypothesize to be the key reason that procedural data enhances pretraining on code and freeform mathematics besides natural language, as we will show in Section~\ref{sec:semantic_understanding}." Weird/confusing to have a speculation (hypothesis) and a lookahead-reference; this box should only be the take-away of the experiments just presented above, not additional speculations.}

%-------------------------------------------------------------------%
\subsection{Where does the Pretrained Information Reside?}
\label{sec:algorithmic_reasoning2}

\textbf{Setup.} We use the \textit{selective transfer} settings defined in Section~\ref{sec:experimental_setup} to probe where useful information is encoded in the pretrained model.
We repeat the experiments from
Section~\ref{sec:algorithmic_reasoning1}
with either \textit{attention-only} or \textit{MLP-only transfer} and compare their performance to full-transfer
to identify which component retains the most benefit. 

\begin{figure*}[t!]
    \centering
    \setlength{\tabcolsep}{1pt}
    \begin{tabularx}{\linewidth}{*{4}{>{\centering\arraybackslash}X}}
        \scriptsize\phantom{m}\textsc{Haystack}&
        \scriptsize\phantom{mm}\textsc{Addition}&
        \scriptsize\phantom{mm}\textsc{Reversed addition}&
        \scriptsize\phantom{mmm}\textsc{Sorting}
    \end{tabularx}\vspace{1pt}
    \includegraphics[trim=0pt 8pt 0pt 32.5pt,clip,width=1.0\linewidth]{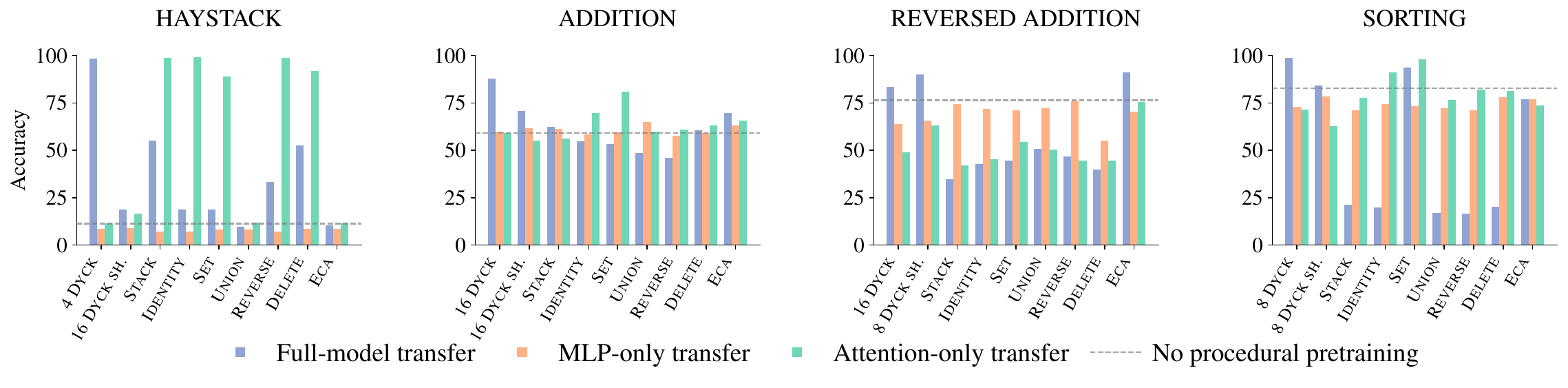}
    \vspace{-12pt}
    \caption{\textbf{Selective transfer of MLP or attention layers
    can improve over full-model transfer},
    % For example, attention layers alone are remarkably useful for \textsc{Haystack}, which involves long-context recall.
    showing that procedural pretraining creates `modular' structure
    localised in the selected model components. Reported values are means across 10 seeds (full results with variance in Appendix~\ref{app:more_results-algorthmic-reasoning}).} 
    %See Appendix~\ref{app:more_results} for full results.}
    \label{fig:alg_reasoning_partial_transfer}
\end{figure*}

\textbf{Results.}
Figure~\ref{fig:alg_reasoning_partial_transfer} shows
surprisingly
that selective transfer can be superior to full-model transfer. 
For instance, with the \textsc{Identity}\,/\,\textsc{Haystack} pair,
attention-only gives an $80$-percentage~point improvement over full-model transfer.
This means that useful information is encoded in the attention layers,
and that the other pretrained components (MLPs) contain non-transferable structure.
Across the different tasks, the attention layers are the most consistent carrier of useful information, with the exception of \textsc{Reversed addition}, where MLP-only and full-model are superior.

\myFrame{\textbf{Take-away.}
Procedural pretraining creates localised skills in specific components of the architecture. Transferring specific components can be more effective than transferring the entire model.}
% \damien{There is no mention of "modularity" or "transferable" in the preceding paragraph. This paragraph is only about locating the useful information.}

%------------------------------%
\subsection{Are There Simple Explanations for the Benefits of Pretraining?}
\label{sec:algorithmic_reasoning3}
% We now test possible mechanisms that could explain how procedural pretraining
% produces the improvements observed in the preceding experiments. 
We next examine potential mechanisms underlying the improvements from procedural pretraining. 
    The goal is to identify whether these gains arise merely from superficial optimisation effects such as generic regularisation or weight rescaling.
See Appendix~\ref{app:simple-explanations} for full details and results.

\textbf{Explanation 1: attention sharpening}. We observe that pretrained models have sharp attention patterns, and transferring only the sharpest attention heads preserves or even exceeds the performance of transferring all of them. 
One possibility is thus that pretraining creates a generic ``sharpening'' of the attention~\citep{liu2023exposing} with no relevance to precise patterns.
However, training models with an explicit regularizer for sharper attentions does not replicate the benefits of procedural pretraining. This shows that precise attention patterns do matter.

\textbf{Explanation 2: initialisation scale.} Another explanation is that pretraining simply adjusts the magnitude of initial weights \citep{huang2020improving,wu2022insights}. 
We test this using the best models from
Section~\ref{sec:algorithmic_reasoning1},
and shuffle the weights per layer, such that the distributions of magnitudes are preserved 
but their structures erased.
% The drop in performance will then indicate how much of the performance is due to this structural information.
% Using 10 seeds, we report the mean results.
% Variance data is in Appendix~\ref{app:more_results}.
As expected, Figure~\ref{fig:algorithmic_shuffled_weights} in the appendix
%Appendix~\ref{??}
shows that the accuracy drops dramatically. %in most cases.
%(except for \textsc{Sorting}, where the benefits are partially preserved). 
We also observe a rapid drop in accuracy with the gradual addition of Gaussian noise to the weights.
% These results reaffirm that the procedurally-pretrained weights encode meaningful structure.
This shows that pretrained weights encode meaningful structure.

% \reb{Together with the improved generalisation on algorithmic tasks, this indicates precise effects induced by procedural data than merely memorisation, better optimisation, or generic attention properties.}

\myFrame{\textbf{Take-away.} The benefits of procedural pretraining are encoded in precise weight structure. They cannot be explained by a simple rescaling of the weights or generic regularisation of the attention.}

%% file: 4_1_semantic_understanding.tex
%\section{Procedural Pretraining on Naturalistic Semantic Data} % D: sounds mysterious
% \section{Combining Procedural Pretraining with Standard Data}
% \section{Is Procedural Data Complementary to Standard Data?}
\section{Can Procedural Data Complement or Replace Standard Data?}
\label{sec:semantic_understanding}

We now examine the \textit{practical} benefits of procedural pretraining.
% on semantic domains.
In Section~\ref{sec:semantic_understanding1},
we use domain-specific datasets (language and code)
%(not mixed with comments in natural language)
to evaluate if the learned \textit{abstract} algorithmic skills (Section~\ref{sec:algorithmic_reasoning})
help in \textit{semantic} domains.
%understand the potential of the transfer. 
In Section~\ref{sec:scaling_up},
we turn to larger pretraining corpora including natural language mixed with code and informal math. 

\subsection{Domain-Specific Corpora}
\label{sec:semantic_understanding1}

\textbf{Setup.}
We use \textsc{WikiText}~\citep{merity2016pointer} and Github's \textsc{JavaCorpus}~\citep{allamanis2013mining} as domain-specific
datasets of natural language and undocumented code. 
We train \textsc{GPT-2}-small models from scratch on these datasets after
initial pretraining on procedural data (full-model transfer).
We repeat this with different amounts of procedural tokens $T_1$
(additive setting).

%We do so by varying the sequence length and the number of pretraining steps.
%\damien{which one? both? why not only adjust the latter?}

\begin{figure}[h!]
\vspace{-5pt}
    \centering
    \includegraphics[width=1\linewidth]{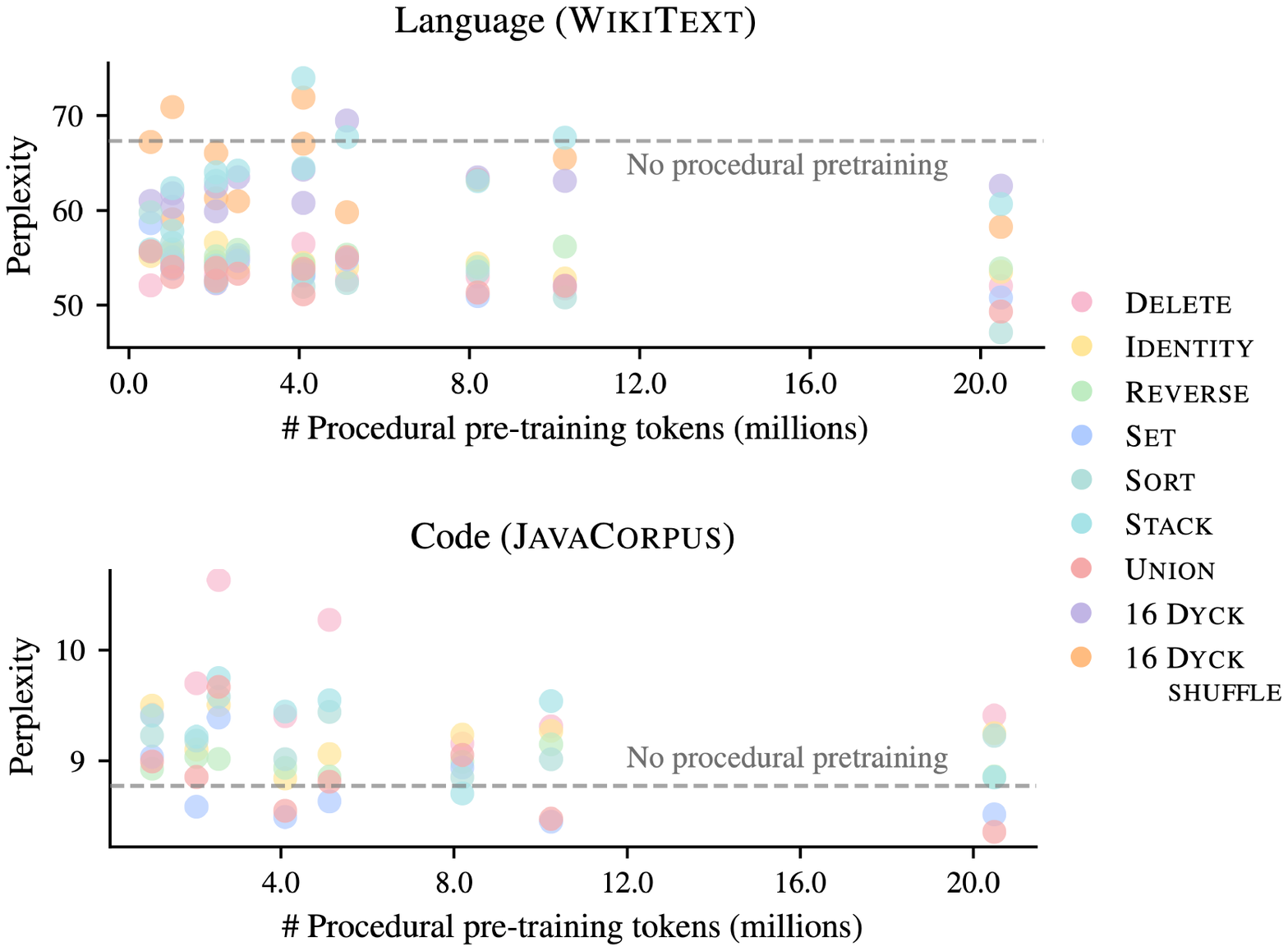}
    \vspace{-12pt}
    \caption{
    \textbf{The benefits of procedural pretraining transfer to semantic domains.}
    Perplexity (lower is better) on natural language
    %(\textsc{WikiText}, \textbf{left})
    (\textbf{left})
    and pure code
    %(\textsc{JavaCorpus}, \textbf{right}).
    (\textbf{right}).
    Introducing a little procedural data is very effective:
    compare the number of procedural tokens ($T_1$) in these plots
    with the amount of tokens from the target datasets
    ($T_2$) being $15$M for \textsc{WikiText}
    and $105$M for \textsc{JavaCorpus}.}
    %, and we do \red{Z} epochs over the data).}
    \label{fig:sem_understanding_results}
    \vspace{-5pt}
\end{figure}

\textbf{Results.}
Figure~\ref{fig:sem_understanding_results} shows that procedural pretraining significantly outperforms the baseline for both natural language and code.
Surprisingly, the improvement is not clearly correlated with the amount of procedural pretraining tokens ($T_1$) and small amounts of pretraining proves sufficient.
%already achieves significant improvements.
Data generated with \textsc{Union} and \textsc{Set} help both domains, while \textsc{Sort}
only helps with natural language.
Additional results in Appendix~\ref{app:grid-search}
show that the sequence length and the number of pretraining steps, both controlling $T_1$, influence the effectiveness of different types of procedural data.
% Much remains to be explained about these various effects.

% \todo{Figure 4, switch from 1*2 to 2*1 plot? (I.e., stack the two subplot vertically and make this plot 1 column)}

\begin{figure*}[t!]
    \centering
    \includegraphics[width=0.9\linewidth]{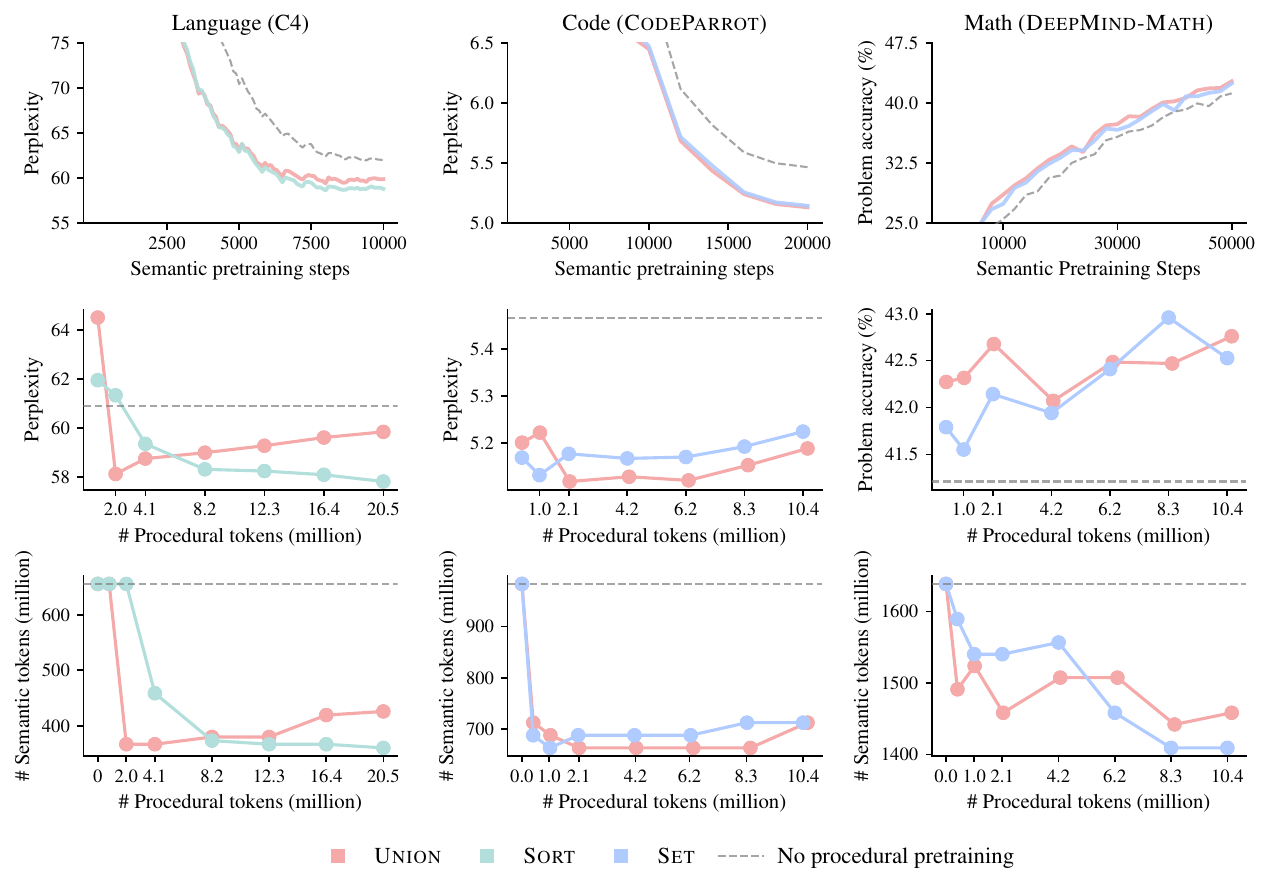}
    \vspace{-8pt}
    \caption{
    \textbf{Procedural pretraining is complementary to standard data and is highly data-efficient.}
    Each column corresponds to a different semantic dataset.
    \textbf{(Top)}~Training curves with different types of procedural data (\textsc{Union}, \textsc{Sort}, \textsc{Set}).
    \textbf{(Middle)}~Additive setting: a small amount of procedural data is sufficient to outperform standard pretraining. %, while additional data does not always yield further gains.
    \textbf{(Bottom)}~Substitutive setting: we plot curves whose points $(x,y)$
    achieve equivalent performance with $x$ procedural tokens and $y$ standard tokens.
    We can drastically reduce the total amount of data when using a small fraction of procedural data. 
    \reb{Full-model transfer (see Section~\ref{sec:experimental_setup}) is used for procedural pretraining.}
    }
    \label{fig:scaling_up_results}
    % \vspace{-9pt}
\end{figure*}

\myFrame{\textbf{Take-away.} The benefits of procedural pretraining transfer from abstract algorithmic skills to semantic domains, and they only require relatively small amounts of data.}

% \subsection{Does Procedural Pretraining Rely Equally on Attention and MLP Layers?}
% \label{sec:semantic_understanding2}

% We established in Section~\ref{sec:algorithmic_reasoning3}
% that different types of procedural data improve different algorithmic skills
% through different components (MLP or attention layers).
% We now investigate whether this is also the case for semantic domains.

% \textbf{Setup.}
% %Building upon the results of , 
% We use the best procedurally-pretrained models
% from Section~\ref{sec:semantic_understanding1}
% and repeat the fine-tuning with selective
% attention-only or MLP-only transfer.
% The difference in performance with full-model transfer then indicates which part of the architecture carries the most useful information.

%% file: 4_2_scaling_up.tex
% \subsection{Are Procedural and Standard Pretraining Complementary?}
\subsection{Larger Pretraining Corpora}
\label{sec:scaling_up}

\textbf{Setup.}
We expand the evaluation to more diverse and larger datasets to evaluate whether 
the knowledge gained from procedural pretraining is complementary to the information typically acquired from these.
We use several standard pretraining datasets for \textbf{natural language}
(\textsc{C4}, \citet{raffel2020exploring}), \textbf{code} (\textsc{CodeParrot}, \citet{codeparrot}), and \textbf{informal mathematics} (\textsc{DeepMind-Math}, \citet{saxton2018analysing}, the math portion of \textsc{The Pile}, \citet{gao2020pile}). 
% \reb{The latter two domains are considered because, 
% % although much prior work (Section~\ref{sec:related_work}) has been limited to natural language, 
% code and mathematics possess strong structural regularities that make them especially well-suited for substantial gains from procedural pretraining.}
\reb{Much of the prior work (see Section~\ref{sec:related_work}) has been limited to natural language, we additionally consider informal mathematics and code because they also constitute an important part of standard pretraining corpora. 
We also hypothesize that they are well-suited for substantial gains from procedural pretraining due to their strong structural regularities similar to procedural data.}

% , to test whether the benefits of procedural pretraining vanish at scale or provide complementary inductive biases under semantic exposure.
% \damien{Unnecessary footnote: "We train models similar to CodeParrot-small from scratch" with a citation in the bibliography of the HF url.}
% \reb{We use the best configurations and types of procedural data (\textsc{Union}, \textsc{Sort},  \textsc{Set}) identified in Figure~\ref{fig:sem_understanding_results} of Section~\ref{sec:semantic_understanding1}.}
% We train models similar to CodeParrot-small~\citep{codeparrot} from scratch.
% Each model is first pretrained on $T_1$ procedural tokens (0--20M)
% then undergoes ``standard'' pretraining on $T_2$ tokens from one of the above datasets (respectively 655M, 1B, or 1.6B tokens).
% We evaluate both the \textit{additive} and \textit{substitutive} settings.
% In the additive case, we measure the absolute performance gain thanks to the $T_1$ procedural tokens.
% In the substitutive case, we assess how many $T_2$ tokens can be saved by $T_1$ tokens without loss of performance. More formally, we measure the savings $\Delta T_2$ such that training on $(T_2\!-\!\Delta T_2)$ semantic tokens with $T_1$ procedural tokens matches the performance of the $T_2$-only model.

\begin{figure*}[t!]
\vspace{-5pt}
    \centering
    \setlength{\tabcolsep}{5pt}
    \begin{tabular}{lc}
    \raisebox{20pt}{\rotatebox{90}{\scriptsize\textbf{\textcolor{midGray}{Small datasets}}}}&
    \includegraphics[width=0.72\linewidth]{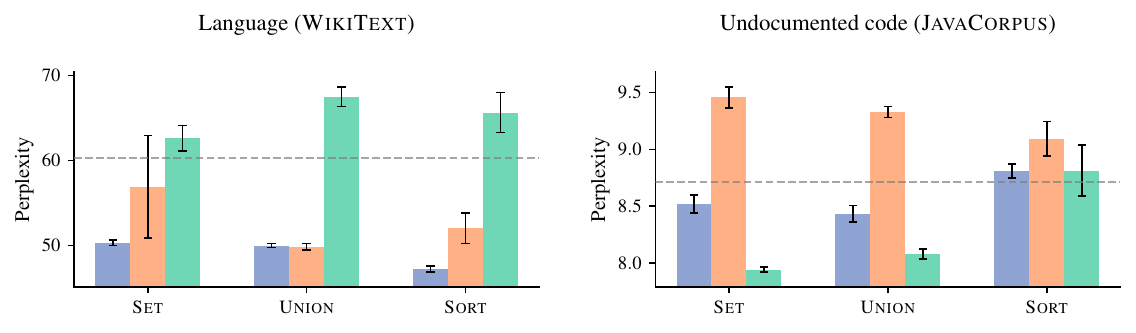}\\
    \raisebox{30pt}{\rotatebox{90}{\scriptsize\textbf{\textcolor{midGray}{Larger datasets}}}}&
    \includegraphics[width=0.9\linewidth]{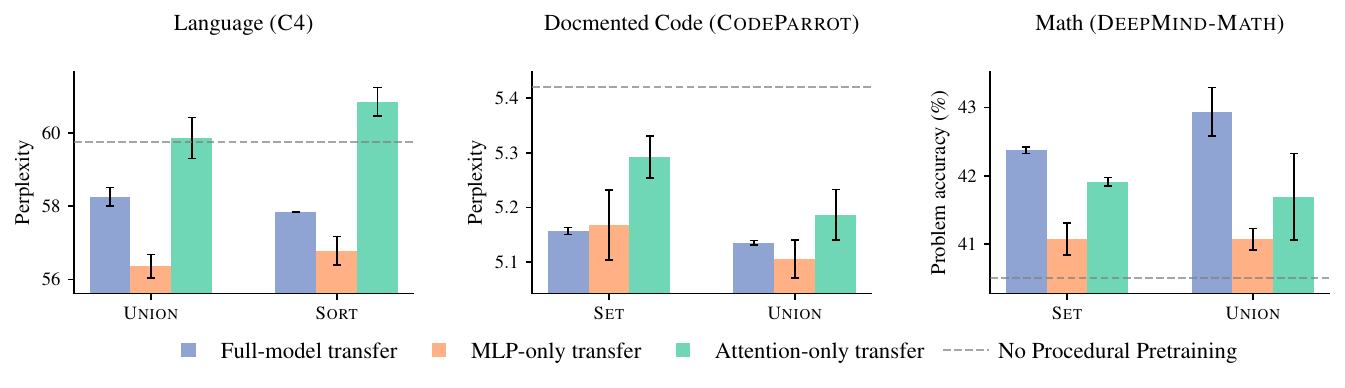}
    \end{tabular}
    \vspace{-7pt}
    \caption{\textbf{Localisation of transferable pretrained information
    for different semantic domains}.
    \textbf{(Top)}
    Using selective weight transfer \reb{described in Section~\ref{sec:experimental_setup}},
    we find that MLPs and attention layers are important respectively for natural language
    and pure code, across different types of procedural data.
    \textbf{(Bottom)}
    On larger datasets,
    MLP-only transfer works best for language.
    As expected, full transfer
    %transferring both attention and MLPs
    is optimal for domains involving both
    language and structured data (documented code, informal mathematics).}
    %larger language corpora, while full-model transfer is the best for domains mixing natural language and structured data.}
    \label{fig:sem_understanding_partial_transfer}
    \vspace{-5pt}
\end{figure*}

\reb{We use the best-performing procedural data types (\textsc{Union}, \textsc{Sort}, \textsc{Set}) identified in Figure~\ref{fig:sem_understanding_results} (Section~\ref{sec:semantic_understanding1}).}
We train CodeParrot-small--style models \citep{codeparrot} from scratch.
Each model is first pretrained on $T_1$ procedural tokens ($0$--$20$M), followed by standard pretraining on $T_2$ tokens from one of the above datasets ($655$M, $1$B, or $1.6$B).
We evaluate both \textit{additive} and \textit{substitutive} settings.
In the additive setting, we measure absolute performance gains from the $T_1$ procedural tokens.
In the substitutive setting, we quantify semantic-token savings $\Delta T_2$ such that training on $(T_2 - \Delta T_2)$ semantic tokens with $T_1$ procedural tokens matches the performance of a $T_2$-only model.

\textbf{Results.} Figure~\ref{fig:scaling_up_results} (top) shows that procedural pretraining accelerates and improves subsequent pretraining.
The {additive setting} (middle) demonstrates that the benefits from procedural pretraining
only require a small amount of data, and that additional data is not always beneficial.
In all cases, a small amount of additional procedural tokens (2--4M) clearly outperform the baseline.
For reference, $2.1$M procedural tokens correspond respectively
to $0.3$\%, $0.2$\%, and $0.1$\% of each of the three semantic datasets.
The {substitutive setting} (bottom) shows that procedural tokens can efficiently substitute for large amounts of semantic tokens.
For example, with \textsc{C4}, we can maintain the baseline loss and save about $45$\% of semantic tokens (and thus FLOPs) by using only $2.1$M procedural tokens. 
For \textsc{CodeParrot} and \textsc{DeepMind-Math}, a $33$\% and $14$\% tokens/FLOPs saving can be achieved with $2.1$M and $10.4$M procedural tokens.

\reb{In Appendix~\ref{app:scaling-procedural-pretraining}, we further examine how the effects of 
procedural pretraining scale with both model size ($350$M and $1.3$B parameter models) and data size (up to $10$B tokens). The larger models 
continue to exhibit clear and consistent improvements from procedural pretraining on a larger scale.}

% Finally, marginal replacement score (MRS) from the substiutive setting analysis shows that procedural tokens can substitute for large amounts of semantic data (Figure~\ref{fig:mrs}). In \textsc{C4}, for example, up to $\sim$60\% of semantic tokens can be replaced without performance loss. Together, these results demonstrate that procedural pretraining not only provides complementary inductive biases but also improves data efficiency at scale.

\myFrame{\textbf{Take-away.} Procedural pretraining is complementary to standard pretraining on semantic datasets in multiple domains.
It is also highly data-efficient and allows one to drastically reduce the total amount of data needed to reach a given perplexity level.}

\subsection{Do the benefits persist on downstream tasks?}
\label{sec:downstream}

We further evaluate if the above benefits of procedural pretraining on the perplexity of standard pretraining transfer to downstream 
tasks, the primary indicator of practical model utility. Following semantic pretraining, 
we evaluate (in fine-tuning or few-shot regimes) both the baseline and our models on representative language 
(WikiText-103~\citep{merity2016pointer}, GLUE~\citep{wang2018glue}), code completion/generation
(PY150~\citep{lu2021codexglue}, MBPP~\citep{austin2021programsynthesislargelanguage}) and commonsense reasoning (ARC-Easy~\citep{clark2018think}, HellaSwag~\citep{zellers2019hellaswag}) datasets. As detailed in Appendix~\ref{app:downstream_finetuning}, the improvements from procedural pretraining consistently persist after downstream fine-tuning. 

\begin{figure}[!h]
% \vspace{-5pt}
    \centering
    \includegraphics[width=0.8\linewidth]{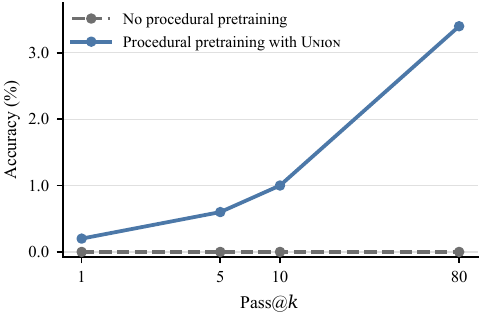}
    \vspace{-2pt}
    \caption{
    \textbf{The benefits of procedural pretraining transfer to downstream tasks.} 
    Here we show $3$-shot pass@$k$ problem accuracy on MBPP. See Appendix~\ref{app:downstream_finetuning} for results on other zero-shot and fine-tuning tasks.}
    %, and we do \red{Z} epochs over the data).}
    \label{fig:mbpp_results}
    \vspace{-5pt}
\end{figure}

Notably, as shown in Figure~\ref{fig:mbpp_results}, a 124M CodeParrot model with procedural pretraining starts to show non-random accuracy on MBPP, while the baseline model shows zero accuracy up to 80 attempts. Again, the compared models are identical in all respects except for the procedural warm-up, which only requires a trivial amount of extra compute. 

These results further support that some of the structure that is imparted from the procedural pretraining is retained throughout subsequent semantic pretraining, echoing the finding of \citet{jesus2021effect}.

\myFrame{\textbf{Take-away.} The benefits of procedural pretraining to standard pretraining persist after downstream fine-tuning.}

%% file: 4_3_partial_transfer_semantic.tex
% \subsection{Layer Specialisation}

\subsection{Localisation of the Transferable Pretrained Information}
\label{sec:partial_transfer_semantic}

% % \damien{Why not setup/results paragraphs like in every section?}
% This section demonstrates that procedural pretraining shows that
% %\textit{layer specialization}, where
% different layers benefit different semantic domains. 
% \todo{L: to be edit}

% We now seek to better understand the mechanisms behind procedural pretraining.
%by locating where the transferable pretrained information resides in the model.

\textbf{Setup.}
Similar to section~\ref{sec:algorithmic_reasoning2}, we use selective weight transfer (attention-/MLP-only)
to locate where the useful, transferable information resides in the procedurally-pretrained model, for the semantic domains considered so far.
%to understand which architectural component contains useful information.
%, comparing attention-only or MLP-only transfer with full-model transfer across different domains. 
Note that we consider \textsc{JavaCorpus} and \textsc{CodeParrot}
as different domains since they respectively contain pure and documented code
(i.e.\ interleaved with natural language).

\textbf{Results.} Figure~\ref{fig:sem_understanding_partial_transfer}
shows that on \textsc{JavaCorpus} (pure code), transferring only the attention layers yields the largest gains in both perplexity and code-completion accuracy (Figure~\ref{fig:java_code_completion_acc}).
On \textsc{WikiText} and \textsc{C4} (natural language), the opposite holds, and transferring the MLPs is more effective than transferring attention layers.
This suggests that procedural pretraining induces distinct inductive biases in different components, \reb{and selectively transferring the right component can further improve upon the results from Figure~\ref{fig:scaling_up_results}}.
For domains that combine natural language with structured data, i.e.\ documented code and informal math (\textsc{CodeParrot} and \textsc{DeepMind-Math}), full-model transfer performs better overall, by combining the benefits from both MLPs for natural language, and attention for structured data.
The fact that transferring MLP is better than attention for natural language is intriguing given that MLPs are believed to
store factual information in LLMs
\citep{dong2025attention,geva2020transformer,xu2025filtering}, raising the question of how procedural pretraining improves MLPs for handling
natural language with only abstract data.

In Appendix~\ref{app:more_results-semantic}, we further explore the benefits of MLP-only transfer for language \reb{on syntactic and morphological competence}. We show that MLP-only transfer achieves a better downstream accuracy on BLiMP~\citep{warstadt2020blimp} in the additive setting. In the substitutive setting, it requires even fewer \textsc{C4} tokens to reach the same perplexity level than full-model transfer ($42$\% vs.\ $55$\%). 

% \begin{figure}[h!]
% \vspace{-5pt}
%     \centering
%     \includegraphics[width=0.9\linewidth]{figures/ppts_partial_transfer_5.pdf}
%     \vspace{-7pt}
%     \caption{Selective transfer across semantic pretraining domains. 
%     For natural language (\textsc{C4}), MLP-only transfer provides the strongest gains. 
%     For code (\textsc{CodeParrot}), both MLP and attention contribute, with full-model transfer slightly stronger. 
%     For mathematics (\textsc{DeepMind-Mathematics}), full-model transfer is clearly best, though each component alone still improves over the baseline.}
%     \label{fig:scaling_up_partial_transfer}
% \end{figure}

% \myFrame{\textbf{Take-away.}
% Procedurally pretrained attention layers are most valuable for structured data such as pure code, whereas MLP layers contribute more to natural language. For larger pretraining corpora that mix both, leveraging both components is essential, making full-model transfer the most effective strategy.
% }

\myFrame{\textbf{Take-away.}
Procedural pretraining instils useful transferable information in both MLPs and attention layers.
MLPs benefit natural language while attention layers support structured domains such as code and mathematics.}

%shows layer specialisation -- benefits differ by component, where attention aids structured data such as code and MLPs aid natural language. For larger pretraining corpora that mix both, leveraging %both components is essential.

%% file: 6_mixtures.tex
\section{Combining Multiple Types of Procedural Data}
\label{sec:mixture}

Our experiments and most prior work on procedural data
so far use a single type of such data at a time.
Combining the strengths of multiple procedural data sources is promising but not trivial due to their varying levels of learning difficulty.
This section explores two techniques to combine the
complementary benefits of multiple types of procedural data
by building on the findings from 
Section~\ref{sec:algorithmic_reasoning}--\ref{sec:semantic_understanding}.

\subsection{Data Mixtures} 

\textbf{Setup.} A natural approach is to pretrain on mixtures of procedural data
in chosen ratios.
%in varying ratios.
We evaluate pairs of procedural data $A$ and $B$
that we mix using $T_A$ and $T_B$ tokens of each, such that $T_1=T_A + T_B$ is fixed.
We prefix each pretraining sequence with an extra token specifying which of $A$ or $B$ it belongs.
We train a model on these $T_1$ tokens then on $T_2$ tokens from either \textsc{JavaCorpus} or \textsc{WikiText}. 

% \begin{figure}[ht!]
% % \vspace{-5pt}
%     \centering
%     \begin{minipage}{0.62\textwidth}
%         \includegraphics[width=0.48\linewidth]{figures/final/wiki_data_mixture_full_transfer_2.pdf}
%         \includegraphics[width=0.48\linewidth]{figures/final/java_data_mixture_attention_transfer_2.pdf}
%     \end{minipage}\hfill
%     \begin{minipage}{0.38\textwidth}
%         \caption{\textbf{Mixtures of two types of procedural data.}
%         We vary the proportion of
%         \textsc{Set} and \textsc{Union}
%         %the two types
%         (indicated by the small pie charts)
%         while keeping the total number of procedural token $T_1$ fixed.
%         %remains fixed while the portions of \textsc{Set} and \textsc{Union} are varied.
%         Some choices achieve a clearly better perplexity \reb{(lower is better)} than either of the two types alone.}
%         %is achievable, showing the potential of procedural data mixture design.}
%         \label{fig:data_mixture_component}
%     \end{minipage}
%     \vspace{-5pt}
% \end{figure}

\begin{figure}[ht!]
    \centering
    \includegraphics[width=0.48\linewidth]{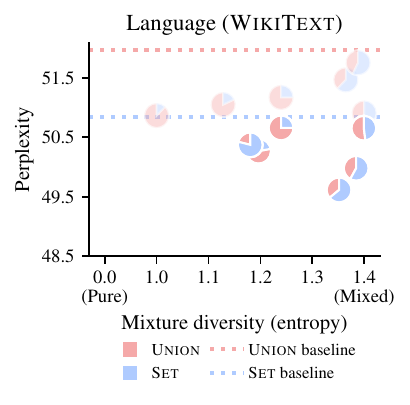}
    % \hfill
    \includegraphics[width=0.48\linewidth]{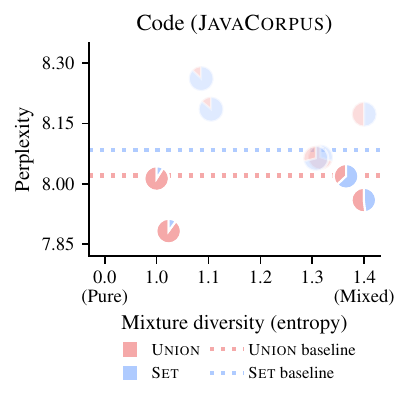}
    \caption{\textbf{Mixtures of two types of procedural data.}
    We vary the proportion of \textsc{Set} and \textsc{Union} (indicated by the small pie charts) while keeping the total number of procedural tokens $T_1$ fixed.
    Some choices achieve clearly better perplexity \reb{(lower is better)} than either type alone.}
    \label{fig:data_mixture_component}
    \vspace{-5pt}
\end{figure}

\textbf{Results.} Figure~\ref{fig:data_mixture_component} shows that many mixtures, \reb{each with different mixture ratios shown by the pie chart and entropy of ratios}, outperform the best single-source baselines for attention transfer on \textsc{JavaCorpus} and full-model transfer on \textsc{WikiText} (the best settings identified in Section~\ref{sec:partial_transfer_semantic}).
This proof of concept
shows that the benefits of multiple types of data are cumulative,
and suggest potential for further gains with optimized combinations of additional sources.

\subsection{Weight Mixtures}
We evaluate an alternative method that builds on the findings from 
Sections~\ref{sec:algorithmic_reasoning3}\,and\,\ref{sec:partial_transfer_semantic}
about the localisation of pretrained information in distinct layers (attention vs.\ MLPs).
We propose to compose a new model by assembling components from several pretrained models.
This avoids the challenge of balancing data mixtures.

\textbf{Setup.} We assemble a model with the attention layers of a pretrained \textsc{Set} model and the MLPs of an \textsc{ECA Rule 110} model. 
We chose these because they showed distinct and complementary capabilities (see \textsc{Haystack} and \textsc{Reversed addition} in Table~\ref{tab:mixture_weights}).
We then further train this model on 
the algorithmic evaluation tasks of Section~\ref{sec:algorithmic_reasoning}.

\textbf{Results.} The last row of Table~\ref{tab:mixture_weights} shows that the combined model yields superior performance across the four tasks,
while the single-source models have weaknesses on one or more tasks.
This indicates that
procedurally-pretrained models
can be modularly combined by simply assembling their most useful components.

% \begin{table}[!h]
% \captionof{table}{\textbf{Pretrained models combined at the weight level.}
% We combine \textsc{Set}-pretrained attention layers with \textsc{ECA}-pretrained MLPs (last row).
% This yields strong performance across all four tasks, whereas single-source models show weaknesses in at least one task. Full results with variance in Table~\ref{tab:mixture_weights_full}.}
% % \vspace{-3pt}
% \centering
% \resizebox{1.0\linewidth}{!}{ % <-- control width here (e.g. 0.85, 0.95, 1.0)
% \begin{tabular}{lcccc}
% \toprule
% & \textsc{Haystack} & \textsc{Addition} & \textsc{Reversed addition} & \textsc{Sort} \\
% \midrule
% \textnormal{No procedural pretraining} & 11.3 & 59.1 & 76.4 & 82.7 \\
% \midrule
% \textsc{Set} (full-model transfer) & 18.9 & 53.4 & 44.6 & 93.5 \\
% \textsc{Set} (attention-only transfer) & \underline{88.9} & \textbf{81.1} & 54.4 & 98.1 \\
% \midrule
% \textsc{ECA} (full-model transfer) & 10.5 & 69.6 & \textbf{91.0} & 76.9 \\
% \textsc{ECA} (MLP-only transfer) & 8.71 & 63.1 & 70.5 & 77.1 \\
% \midrule
% \textsc{Set} (attention) + ECA (MLP) & \textbf{94.4} & \underline{80.3} & \underline{82.9} & \textbf{99.4} \\
% \bottomrule
% \end{tabular}
% }
% \label{tab:mixture_weights}
% \vspace{-5pt}
% \end{table}

\begin{table}[!h]
\captionof{table}{\textbf{Procedurally-pretrained models combined at the weight level.}
We combine \textsc{Set}-pretrained attention layers with \textsc{ECA}-pretrained MLPs (last row).
This yields strong performance across all four tasks, whereas single-source models show weaknesses in at least one task. Full results with variance in Table~\ref{tab:mixture_weights_full}.}
\centering
\resizebox{1.0\linewidth}{!}{
\begin{tabular}{lcccc|c}
\toprule
& \textsc{Haystack} & \textsc{Addition} & \makecell{\textsc{Reversed} \\ \textsc{addition}} & \textsc{Sort} & \textsc{Avg.} \\
\midrule
\textnormal{No procedural pretraining}
& 11.3 & 59.1 & 76.4 & 82.7 & 57.4 \\
\midrule
\textsc{Set} (full-model transfer)
& 18.9 & 53.4 & 44.6 & 93.5 & 52.6 \\
\textsc{Set} (attention-only transfer)
& \underline{88.9} & \textbf{81.1} & 54.4 & 98.1 & \underline{80.6} \\
\midrule
\textsc{ECA} (full-model transfer)
& 10.5 & 69.6 & \textbf{91.0} & 76.9 & 62.0 \\
\textsc{ECA} (MLP-only transfer)
& 8.71 & 63.1 & 70.5 & 77.1 & 54.9 \\
\midrule
\textsc{Set} (attention) + \textsc{ECA} (MLP)
& \textbf{94.4} & \underline{80.3} & \underline{82.9} & \textbf{99.4} & \textbf{89.3} \\
\bottomrule
\end{tabular}
}
\label{tab:mixture_weights}
\vspace{-8pt}
\end{table}

\myFrame{\textbf{Take-away.} The effects of multiple types of procedural data
are additive. Proof-of-concept experiments show that they can be combined both at data and weight levels, and suggest ample room for further benefits with larger and more-optimised combinations.}
% \vspace{6pt}

%% file: 8_conclusion.tex
\section{Discussion}
\label{sec:discussion}
This paper shows that pretraining language models on
well-chosen abstract procedural data complements 
standard pretraining, accelerating training and improving performance
on natural language, code, and informal mathematics. 
Our experiments also shed light on the origin of these gains. We found that
useful information lies in different components
(MLP vs.\ attention)
depending on the domain
(language vs.\ structured domains).
\reb{These findings motivate new pretraining paradigms where primitive abstract data is exposed to LLMs before they acquire rich world knowledge.}
% and undergo further downstream finetuning.}
% Explaining the variety of benefits
% from different types of procedural data
% is an important question for future work.

% \vspace{3pt}
\textbf{Efficient initialisation.}
%Procedural pretraining is computationally expensive although the 
Unlike standard data, procedural data has a small Kolmogorov complexity,
meaning that it contains information that can be summarized in 
%since its generation rests on
a few lines of code.
% Generating millions of samples processed by gradient descent thus seems computationally wasteful. 
In principle, it may be possible to simplify this as a deterministic or closed-form \emph{smart initialisation} of LLMs.

\reb{\textbf{Why is procedural data helpful?} 
% A first-principles explanation of why specific forms of data help is a promising future direction. 
Our results in Section~\ref{sec:algorithmic_reasoning3} rule out simple explanations, indicating deeper effects than merely a better optimisation dynamics or memorisation. 
Investigating a first-principles explanation or studying the mechanisms at play 
using mechanistic interpretability techniques~\citep{conmy2023automated} is a promising avenue.}

\textbf{Combining multiple types of procedural data.}
We showed that the benefits can be additive. Existing methods for
data mixture optimization \citep{fan2023doge,xie2023doremi,xie2025chameleon}
could be adapted to optimally balance multiple types of procedural data.

% \vspace{3pt}
\textbf{Knowledge vs.\ reasoning.}
\citet{han2025position} argue that LLMs' limitations stem from
entangled representations of knowledge and reasoning.
Our work can be viewed as injecting an `algorithmic reasoning prior' before world-knowledge acquisition.
% This may ultimately provide a mechanism to help disentangle the acquisition of knowledge from reasoning. %in LLMs.
This ultimately suggests a data-driven path in improving knowledge and reasoning acquisition beyond architectural changes~\citep{pouransari2025pretraining}.

% \vspace{3pt}
\textbf{Limitations.}
(1)~We use smaller models (up to $1.3$B) and lower data-model ratio compared to state-of-the-art LLMs, \reb{further} scaling up our experiments is an important future step.
% (2)~Our evaluation is limited to perplexity and accuracy on common benchmarks.
% \reb{\st{The downstream performance of the models has yet to be studied.}}
(2)~While our experiments on combining multiple types of procedural data are a proof of concept, they lay out several promising directions for future work.

%% file: 9_appendix.tex
\section*{Appendix}
The appendix provides the following additional details and results:

\begin{itemize}[itemsep=-2pt,topsep=-2pt,leftmargin=11pt]
    \item Appendix~\ref{app:related_work}: extended review of the related literature.
    \item Appendix~\ref{app:procedural-pretraining}: details about procedural pretraining. 
    \item Appendix~\ref{app:model_details}: details about models used in experiments.
    \item Appendix~\ref{app:algorithmic_task_details}: implementation details for the algorithmic downstream tasks.
    \item Appendix~\ref{app:experimental-details}: training details including hyperparameters for each experiment.
    \item Appendix~\ref{app:simple-explanations}: testing simpler explanations for procedural pretraining benefits.
    \item Appendix~\ref{app:grid-search}: investigates sequence length and number of steps for procedural pretraining. 
    \item Appendix~\ref{app:longer_sequences}: examines longer sequence lengths during procedural pretraining.
    \item Appendix~\ref{app:transferability_analysis}: analysis of the relationship between procedural pretraining loss and downstream semantic performance. 
    \item Appendix~\ref{app:vocab_size}: analyses the impact of vocabulary size during procedural pretraining. 
    \item Appendix~\ref{app:weight_decay}: study of weight decay during procedural pretraining. 
    \item \reb{Appendix~\ref{app:scaling-procedural-pretraining}: study of procedural pretraining with scaling model and semantic dataset size.}
    \item {\reb{Appendix~\ref{app:downstream_finetuning}: evaluates effects of procedural pretraining after downstream fine-tuning.}}
    \item Appendix~\ref{app:more_results}: additional and full results.
\end{itemize}

\input{9_related_work_long}

\section{Procedural Pretraining}
\label{app:procedural-pretraining}

\begin{figure}[h!]
    \centering
    % Left: table
    \begin{minipage}[t]{0.4\linewidth}
        \vspace{0pt}
        \centering
        \scriptsize
        \begin{tabular}{ll}
            \toprule
            Pretraining task & Example sequence \\
            \midrule
            \textsc{$k$-Dyck} &
            \texttt{(} \blue{\texttt{[}} \orange{\texttt{\{}} \orange{\texttt{\}}} \blue{\texttt{]}} \texttt{)} \\
            \textsc{$k$-Dyck Shuffle} &
            \texttt{(} \blue{\texttt{[}} \orange{\texttt{\{}} \blue{\texttt{]}} \texttt{)} \orange{\texttt{\}}} \\
            \textsc{Stack} &
            \texttt{1} \texttt{2} \texttt{3} \texttt{P} \texttt{|} \textbf{\texttt{2}} \textbf{\texttt{1}} \\
            \textsc{Identity} &
            \texttt{1} \texttt{2} \texttt{3} \texttt{|} \textbf{\texttt{1}} \textbf{\texttt{2}} \textbf{\texttt{3}} \\
            \textsc{Set} &
            \texttt{1} \texttt{2} \texttt{2} \texttt{|} \textbf{\texttt{1}} \textbf{\texttt{2}} \\
            \textsc{Sort} &
            \texttt{3} \texttt{1} \texttt{2} \texttt{|} \textbf{\texttt{1}} \textbf{\texttt{2}} \textbf{\texttt{3}} \\
            \textsc{Reverse} &
            \texttt{1} \texttt{2} \texttt{3} \texttt{|} \textbf{\texttt{3}} \textbf{\texttt{2}} \textbf{\texttt{1}} \\
            \textsc{Union} &
            \texttt{1} \texttt{2} \texttt{|} \texttt{2} \texttt{3} \texttt{|} \textbf{\texttt{1}} \textbf{\texttt{2}} \textbf{\texttt{3}} \\
            \textsc{Delete} &
            \texttt{1} \texttt{2} \texttt{3} \texttt{|} \texttt{2} \texttt{|} \textbf{\texttt{1}} \textbf{\texttt{3}} \\
            \bottomrule
        \end{tabular}
    \end{minipage}%  
    % Right: image
    \begin{minipage}[t]{0.24\linewidth}
        \vspace{0pt}
        \centering
        \includegraphics[width=0.75\linewidth]{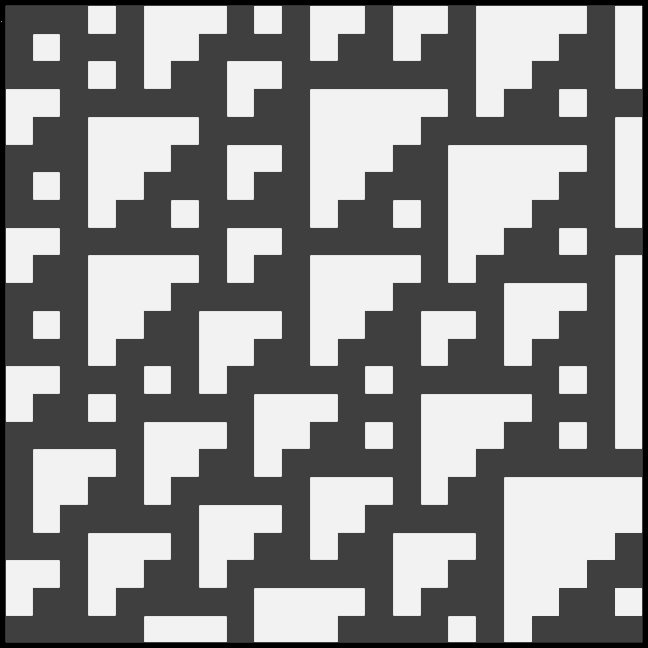}
    \end{minipage}
    \caption{
        We pretrain transformers on various forms of procedural data generated from simple algorithms, such as formal languages (left) or elementary cellular automata (right). In \textsc{$k$-Dyck} examples, matching brackets are color-coded. For \textsc{Stack}, `\texttt{P}' denotes the \textit{pop} operation. The symbol `\texttt{|}' acts as a delimiter between the input and the expected output, on which the loss is computed (\textbf{\texttt{bold}} tokens). For \textsc{Union} and \textsc{Delete}, the first delimiter separates the two sequences to which the transformation is applied, and the second delimiter separates the entire input from the target output.
    }
    \label{fig:procedural_data_examples}
\end{figure}

% \todo{L: maybe it would be nice (if we have time), to include one or two examples of python-style pseudo-code on how the data is generated}

\textbf{Sequence Transformations and Memory Operations Input Sequence Lengths.}
 
For the sequence transformation and memory operation tasks in Section~\ref{sec:algorithmic_reasoning}, procedural pretraining follows a curriculum learning scheme: models begin with input sequences of length 2 or 4 (depending on the task), and the length is increased by 2 once 99\% accuracy is achieved, continuing until a maximum length of 20.  

In Section~\ref{sec:semantic_understanding}, larger transformers are instead pretrained on procedural tasks with fixed input lengths of 8, 16, 32, and 64. Appendix~\ref{app:grid-search} analyses the effect of sequence length, while Appendix~\ref{app:longer_sequences} examines the impact of extending lengths further.

For consistency in token counts, we assume the output sequence is at most twice the length of the input, and thus estimate and report the total number of procedural tokens as $2\times$ the input length.

\textbf{Sequence Transformation Descriptions.}

\textsc{Identity.} The input is a sequence of tokens followed by a separator.  
The target is an exact copy of the input sequence.  
The vocabulary has 102 tokens: 100 valid elements, one separator, and one padding token.  

\vspace{4pt}
\textsc{Set.} The input is a sequence of tokens followed by a separator.  
The target is the same sequence with duplicates removed, preserving the order of first appearance.  
The vocabulary has 102 tokens: 100 valid elements, one separator, and one padding token.  

\vspace{4pt}
\textsc{Union.} The input consists of two token sequences separated by a delimiter.  
The target is the union of both sequences, preserving the order of first appearance.  
The vocabulary has 103 tokens: 100 valid elements, one separator, one padding token, and one union delimiter.

\vspace{4pt}
\textsc{Delete.} The input is a sequence of tokens followed by a separator and a designated token.  
The target is the sequence with all instances of the designated token removed.  
The vocabulary has 103 tokens: 100 valid elements, one separator, one padding token, and one delete marker.

\vspace{4pt}
\textsc{Sort.} The input is a random sequence of tokens followed by a separator.  
The target is the same sequence sorted in ascending numerical order.  
The vocabulary has 102 tokens: 100 valid elements, one separator, and one padding token.

\vspace{4pt}
\textsc{Reverse.} The input is a sequence of tokens followed by a separator.  
The target is the same sequence in reverse order.  
The vocabulary has 102 tokens: 100 valid elements, one separator, and one padding token.

\textbf{Memory Operation Descriptions.}

\textsc{Stack.} The input encodes a sequence of \texttt{push} and \texttt{pop} operations, followed by a separator. The target is the final stack contents, listed top-to-bottom. Tokens are pushed with 75\% probability in the first two thirds of the input and popped with 75\% probability in the final third. Each push inserts a unique token, pops remove the top element, and only one copy of a token may exist on the stack at any time.  The vocabulary has 103 tokens: 100 pushable elements, one \texttt{pop} token, one separator, and one padding token.

\textbf{Other Procedural Data Source Descriptions.}

\textsc{$k$-Dyck.} We generate sequences of correctly nested parentheses using $k$ distinct bracket pairs (vocabulary size $2k$), with $k \in \{4, 8, 16\}$. 
All training sequences are fixed to length 128 and constructed incrementally via a stack-based procedure ensuring syntactic validity. 
At each step, the generator samples an opening or closing bracket with probability $p_{\text{open}}=0.49$ \citep{papadimitriou2023injecting}, 
forcing closure when the remaining token budget matches the number of open brackets.  

%\item 
\vspace{4pt}
\textsc{$k$-Dyck shuffle.} This variant retains the same $2k$-token vocabulary of bracket pairs but removes the requirement of proper nesting. 
Sequences are sampled with a 50\% probability of opening brackets and fixed to length 128, with $k \in \{4, 8, 16\}$. 
While every opening bracket is eventually closed, truncation can yield ill-formed strings \citep{hu2025between}, 
though we did not observe adverse effects in practice.

\vspace{4pt}
\textsc{ECA Rule 110.} We follow the setup of \citet{zhang2024intelligence}, generating data from Elementary Cellular Automata under Rule~110, a Class IV system with Turing-complete dynamics. 
To model binary state sequences with \textsc{GPT-2}, the embedding layer is replaced by a linear projection from binary vectors, and the output softmax is replaced by a projection back to binary space, preserving determinism. 
For transfer, we average the learned input embeddings across the ECA data and use this representation to initialize the embedding layers of downstream transformers.

\section{Model details}
\label{app:model_details}

We use a \textsc{GPT-2}-type architecture~\citep{radford2019language} throughout our experiments.  
In Section~\ref{sec:algorithmic_reasoning}, we employ a minimal configuration with 2 layers, 4 attention heads, and a hidden size of 16 for \textsc{Haystack}, \textsc{Addition}, \textsc{Reversed addition} and \textsc{Sorting}. For \textsc{Multiplication}, we use a model size of 4 layers, 8 attention heads and a hidden size of 512.
In Section~\ref{sec:semantic_understanding} and \ref{sec:mixture}, we use the \texttt{small} \textsc{GPT-2} variant with 12 layers, 12 attention heads, and a hidden dimension of 768.  

\section{Algorithmic Task Descriptions}
\label{app:algorithmic_task_details}

\textbf{Memory Recall.}

\textsc{Haystack.} 
This task tests a model’s ability to retrieve information from long sequences. 
Each input consists of a sequence of key--value pairs of the form 
\([m_1, c_1, m_2, c_2, \ldots, m_k, c_k, m_u]\), 
where each \(m_i\) is a unique marker and \(c_i\) its associated value. 
The sequence terminates with a query marker \(m_u\), and the model must locate its earlier occurrence in the context and output the corresponding value \(c_u\). 
We fix \(k=30\) in all experiments and report accuracy based on whether the predicted value matches \(c_u\).  

\textbf{Arithmetic.}

\textsc{Addition.}
This task probes a model’s ability to learn the compositional structure of arithmetic addition when expressed in \emph{forward} (non-reversed) notation. In this setting, the least significant digits, crucial for carry operations, appear at the \emph{end} of the sequence. As a result, transformers must propagate carry information \emph{backward} through the context, a dependency pattern misaligned with the autoregressive training objective. Each input takes the form \texttt{a+b=}, where \(a\) and \(b\) are randomly sampled \(n\)-digit integers. Inputs and outputs are digit-tokenized, with operator symbols (\texttt{+}, \texttt{=}) assigned unique tokens. The model is trained to predict only the result digits, and cross-entropy loss is computed solely on these positions. For all experiments we fix \(n=5\), and report token-level accuracy on the predicted sum.  

\textsc{Reversed addition.}  This variant evaluates the same underlying arithmetic skill as \textsc{Addition}, but aligns the sequence structure with the autoregression of the transformer. Both input and output sequences are reversed, so carry propagation proceeds left-to-right in the same direction as generation. For example, the sum \(ab + cd = efg\) is represented as input \texttt{b a d c} with output \texttt{g f e}. The task reduces long-range dependencies while preserving the need for multi-step reasoning. We set \(n=10\) and evaluate using token-level accuracy.  

\textsc{Multiplication.}  
This task evaluates a model’s ability to perform multi-digit multiplication. 
Each input takes the form \(a \times b =\), where \(a\) and \(b\) are randomly sampled \(n\)-digit integers. 
The model must generate the digit sequence corresponding to their product. 
Inputs and outputs are tokenized at the digit level, with the multiplication operator (\texttt{\texttimes}) and equals sign (\texttt{=}) assigned special tokens. 
For all experiments we fix \(n=5\). 
Cross-entropy loss and token-level accuracy are computed only on the output positions corresponding to the product digits.  

\textbf{Logical and relational processing.}

\textsc{Sorting.}
This task assesses a model's ability to perform algorithmic reasoning by sorting a sequence of integers. 
Each input consists of a list of \(n\) integers sampled uniformly from the range \([0, P-1]\), where \(P\) denotes the vocabulary size. 
We fix \(n=10\) and \(P=100\). 
The input sequence is followed by a separator token, after which the model must output the sorted version of the sequence. 
For example, the input \texttt{6 3 5 |} requires the output \texttt{3 5 6}. 
Training is autoregressive, and evaluation is performed only on the output positions following the separator, with token-level accuracy as the metric.

\section{Experimental Details}
\label{app:experimental-details}
\subsection{Procedural Pretraining}
\label{app:experimental-details_procedural-pretraining}
\textbf{Details for Section~\ref{sec:algorithmic_reasoning}.}

The hyperparameters used for procedural pretraining are summarised in Table~\ref{tab:pretraining-config}, with the exception of \textsc{ECA Rule 110}, whose configuration is reported separately below.

\begin{table}[h!]
\vspace{-3pt}
\begin{center}
\begin{small}
\begin{sc}
\begin{tabular}{lccc}
\toprule
\textnormal{Task} & Seq. length & Learning rate & Vocab. size \\
\midrule
\textsc{Identity}       & 4–20    & $5 \times 10^{-4}$ & 102 \\
\textsc{Set}            & 2–20    & $5 \times 10^{-4}$ & 102 \\
\textsc{Stack}          & 4–20    & $5 \times 10^{-4}$ & 103 \\
\midrule
\textsc{$k$-Dyck}       & 128     & $5 \times 10^{-5}$ & $2 \times k$ \\
\textsc{$k$-Dyck Shuffle} & 128   & $5 \times 10^{-5}$ & $2 \times k$ \\
\bottomrule
\end{tabular}
\end{sc}
\end{small}
\end{center}
% \vspace{-7pt}
\captionof{table}{
Pretraining hyperparameters for each procedural task. All models use AdamW with weight decay $0.01$, batch size $256$, and run for $1{,}000{,}000$ steps. Early stopping (100 validation checks) is applied for the algorithmic tasks. }
\label{tab:pretraining-config}
\vspace{-2pt}
\end{table}
% \todo{Double check learning rate used for Dyck Languages.}

\textsc{ECA Rule 110.} Following \citet{zhang2024intelligence}, we pretrain models on data procedurally generated from Elementary Cellular Automata under Rule~110. 
Each epoch begins from a new random initial state, ensuring continual access to fresh samples and effectively unlimited training data. 
Models are trained for up to 10{,}000 epochs with early stopping on validation loss. 
We use Adam with a learning rate of $2 \times 10^{-6}$, weight decay $0.01$, and gradient clipping at norm $1.0$, with batch size 64 (60 time steps, 100 spatial dimensions). 
The learning rate schedule consists of a 10\% warm-up phase followed by cosine decay.  

\textbf{Detail for Section~\ref{sec:semantic_understanding}.}

For all algorithmic procedural tasks used in this section (\textsc{Identity}, \textsc{Set}, \textsc{Union}, \textsc{Delete}, \textsc{Sort}, \textsc{Reverse}, and \textsc{Stack}), 
we train using AdamW with a batch size of 64 and no warmup steps. 
Following \citet{hu2025between}, we pretrain models on procedural data with a weight decay of 0.1 for \textsc{Wikitext} and \textsc{C4}, and use 0.01 for \textsc{JavaCorpus}, \textsc{CodeParrot}, and \textsc{DeepMind-Math}. The pretrained models are subsequently fine-tuned on their respective downstream datasets.
An ablation study in Appendix~\ref{app:weight_decay} confirms that this choice of weight decay during pretraining does not affect our conclusions. 
We sweep sequence lengths over \(\{8, 16, 32, 64\}\) and vary the number of procedural pretraining steps between 100 and 2500. 
No warmup or learning rate decay is applied; instead, we train with a fixed learning rate throughout. 
For consistency, the learning rate during pretraining is matched to that of the downstream semantic objective, as preliminary experiments indicated this setting to be most effective.

\subsection{Algorithmic Tasks}
\label{app:experimental-details_algorithmic-tasks}
\textsc{Haystack}, \textsc{Forward addition}, \textsc{Reversed addition}, and \textsc{Sorting}.~
We trained models for $10^4$ steps with a batch size of 1,000. The training data is generated dynamically. We used the AdamW optimizer with a learning rate of $10^{-3}$ and weight decay of $10^{-3}$. We always use an architecture consisting of 2 layers, 4 attention heads, and 16-dimensional embeddings. We report mean and standard deviation over 10 seeds in Appendix~\ref{app:more_results}.

\textsc{Multiplication}.~
These experiments employed a larger model with 4 layers, 8 attention heads, and 512-dimensional embeddings. Thus, we use a smaller training batch size (64 vs. 1,000), resulting in approximately 156k update steps compared to 10k steps for the afforementioned reasoning tasks, despite using the same number of training examples. We optimize with AdamW using a learning rate of $10^{-3}$, weight decay of $10^{-3}$, and 500 warmup steps. We run this over 3 seeds, and report standard deviations in Appendix~\ref{app:more_results}.

\subsection{Semantic Data}

\textsc{WikiText.}
We train our models on {Wikitext-2}~\citep{merity2016pointer} using next-token prediction with AdamW. Training runs for $\sim$7 epochs (5{,}000 steps) with an effective batch size of 32. We use a learning rate of $5 \times 10^{-4}$ with cosine decay and no warmup steps.
Sequences are tokenized with the \textsc{GPT-2} tokenizer, truncated to 1{,}024 tokens. We evaluate the model on the validation split, using 1{,}024 samples. 
Our primary metric is validation perplexity. 

\textsc{JavaCorpus:}
We train our models on Github's {JavaCorpus}~\citep{allamanis2013mining} using next-token prediction with AdamW. Training runs for 5 epochs with an effective batch size of 8. We use a learning rate of $8 \times 10^{-5}$ and no warmup steps. 
The hyperparameters follow those in~\citep{lu2021codexglue}.
Sequences are tokenized with the CodeGPT~\citep{lu2021codexglue} tokenizer, with block size 1{,}024 tokens. We report validation perplexity and test accuracy for code completion.

\textsc{C4:} We pretrain our models on the {C4} dataset~\citep{raffel2020exploring} using next-token prediction with AdamW. 
Training runs for 10{,}000 steps with an effective batch size of 32. 
We use a learning rate of $5 \times 10^{-4}$ with cosine decay and no warmup steps. 
Sequences are tokenized with the \textsc{GPT-2} tokenizer and truncated to 2{,}048 tokens. 
We evaluate models on the C4 validation split using 1{,}024 samples, reporting validation perplexity. 
To assess linguistic generalization, we also report accuracy on the BLiMP grammaticality judgment benchmark \citep{warstadt2020blimp}, which tests whether models prefer grammatical over ungrammatical sentence pairs.

\textsc{CodeParrot:} We pretrain our models on the {CodeParrot} dataset\footnote{\url{https://huggingface.co/datasets/codeparrot/codeparrot-clean}} using next-token prediction with AdamW. 
Training runs for 20{,}000 steps with an effective batch size of 48. 
We use a learning rate of $5 \times 10^{-4}$ with cosine decay, no warmup steps, and weight decay of $0.1$. 
Sequences are tokenized with the CodeParrot's tokenizer and with length 1{,}024 tokens. 
We evaluate models on the validation split with 1{,}000 evaluation steps and a batch size of 48, reporting validation loss and perplexity. 

\textsc{Deepmind-Math:} We pretrain our models on the {Deepmind-Mathematics} dataset \citep{saxton2018analysing} using next-token prediction with AdamW. 
Training runs for 50{,}000 steps with an effective batch size of 64. 
We use a constant learning rate of $8 \times 10^{-5}$ (as is done in the original paper), no warmup steps, and weight decay of $0.1$. 
Sequences are tokenized at the character-level (including digits, alphabet in upper and lower case, punctuation and whitespace, a total of 95 different tokens) and have a length 512 tokens. 
We evaluate models on the in-distribution validation split with 100 evaluation steps and a batch size of 64, reporting the accuracy on the validation problems. 
This ensures evaluating around 38{,}000 questions in each validation session.
A problem is considered correct if and only if the prediction exactly matches the groundtruth answer.

For the 124M-parameter models trained on C4 and \textsc{CodeParrot}, we follow the training recipe of HuggingFace's CodeParrot-small model and the setup used by~\citet{hu2025between}. For experiments on \textsc{DeepMind-Math}, we additionally follow the settings from the original dataset paper~\citep{saxton2018analysing}. For the larger 350M- and 1.3B-parameter models, we use the architectures and learning rates from the EleutherAI Pythia suite~\citep{biderman2023pythia}.

\subsection{Downstream Finetuning}
\label{app:downstream_setup}

\reb{
\textsc{Wikitext-103:} We finetune our language models on the \textsc{Wikitext-103} dataset~\citep{merity2016pointer}. Finetuning runs for $\sim$37 million tokens (10,000 steps) with an effective batch size of 32. We use a learning rate of $1 \times 10^{-4}$ with cosine decay and no warmup steps. Sequences are tokenized with
the GPT-2 tokenizer, truncated to 2,048 tokens.}

\reb{
\textsc{GLUE:} We finetune our language models on the GLUE benchmark~\citep{wang2018glue}. For all evalautions, fine-tuning is run for one epoch with a batch size of 16 and learning rate of $5\times10^{-5}$ with a linear decay.}

\reb{
\textsc{Py150:} We finetune our models on PY150~\citep{raychev2016probabilistic}, which is an influential task evaluating code completion capability~\citep{lu2021codexglue}. 
It contains 150{,}000 Python source files collected from GitHub. 
We first follow~\citet{lu2021codexglue} for the preprocessing and then finetune the models using next-token prediction with AdamW. 
Training runs for 2 epochs with an effective batch size of 8, a learning rate of $8 \times 10^{-5}$, and a 0.01 weight decay. 
Sequences are tokenized with the CodeGPT~\citep{lu2021codexglue} tokenizer, with block size 1{,}024 tokens. We report test accuracy (token-level accuracy) on this task.
}

\section{Testing Simpler Explanations}
\label{app:simple-explanations}

\subsection{Attention Sharpening}
\label{app:attention-sharpening}
\input{9_procedural_attention_analysis}

\subsection{Weight Scaling}
\label{app:weight-shuffling}
We test whether the benefits of procedural pretraining arise solely from weight distribution adjustments, as opposed to precise weight structures and values. Our results show that the gains depend critically on the latter.
% Can the benefits of procedural pretraining be attributed solely to adjustments in the weight distribution, rather than to precise weight structures and values? Our results demonstrate that the gains depend critically on the latter.

\begin{figure}[H]
    \centering
        \begin{minipage}[t]{0.48\linewidth}
        \centering
        \includegraphics[width=\linewidth]{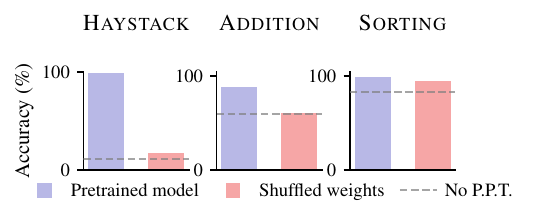}
        \caption{Layer-wise weight shuffling largely eliminates the benefits of procedural pretraining, despite preserving the overall distribution of weight values. This indicates that the advantages arise from precise structural organisation of the weights, rather than from their distribution alone.}
        \label{fig:algorithmic_shuffled_weights}
    \end{minipage}\hfill
    \begin{minipage}[t]{0.48\linewidth}
        \centering
        \includegraphics[width=\linewidth]{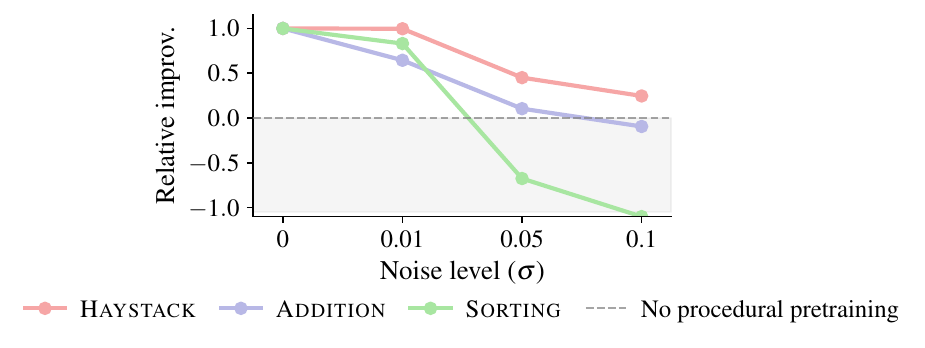}
        \caption{Injecting Gaussian noise into pretrained weights progressively erodes the benefits of procedural pretraining. This demonstrates that precise weight values are essential, and coarse statistics such as weight magnitudes alone cannot account for the performance benefits.}
        \label{fig:noise_injection}
    \end{minipage}

\end{figure}

\textbf{Weight Shuffling.} We apply layer-wise shuffling of the pretrained weights to the best-performing models from Section~\ref{sec:algorithmic_reasoning2} and evaluate downstream accuracy after fine-tuning. This setup explicitly preserves weight distributions while erasing structure.  Figure~\ref{fig:algorithmic_shuffled_weights} demonstrates that weight distributions alone are insufficient: performance collapses to the no procedural pretraining baseline, except for \textsc{Sorting}, which retains partial benefits. We use 10 seeds and report mean results, with variance data in Appendix~\ref{app:more_results}.

\textbf{Noise Injection.}
We introduce additive Gaussian noise to the procedurally pretrained weights of the best models from Section~\ref{sec:algorithmic_reasoning2} and evaluate performance after fine-tuning. We report a relative improvement score, where 1.0 corresponds to unperturbed pretrained weights and 0.0 corresponds to a baseline without procedural pretraining (random initialisation). Figure~\ref{fig:noise_injection} shows that gradually increasing Gaussian noise consistently degrades performance, confirming that precise weight values are crucial. We use 10 seeds and report mean results, with variance data in Appendix~\ref{app:more_results}.

\section{Procedural Data Hyperparameter Grid Search}
\label{app:grid-search}
We study the influence of both pretraining steps and input sequence length on the effectiveness of procedural pretraining for downstream semantic tasks.

\paragraph{Setup.} 
We conduct a grid search over sequence length and number of pretraining steps to determine which configurations of procedural pretraining yield the lowest semantic validation perplexity. 
Each model is first pretrained on a single procedural task for \(T_1\) tokens, followed by \(T_2\) tokens of semantic data (\textsc{WikiText} for natural language and \textsc{JavaCorpus} for code), with full-model transfer. 
The value of \(T_1\) is varied by adjusting the sequence length and number of pretraining steps, while \(T_2\) remains fixed.  

\begin{figure}[h!]
    \centering
    \includegraphics[width=1.0\linewidth]{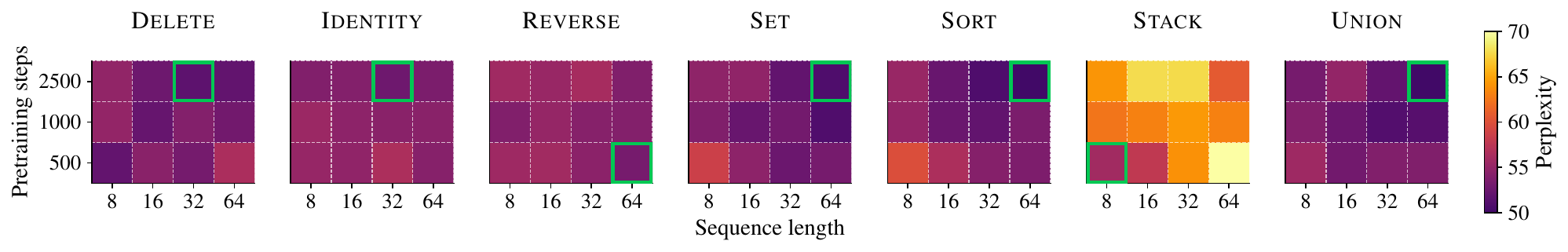}
    \includegraphics[width=1.0\linewidth]{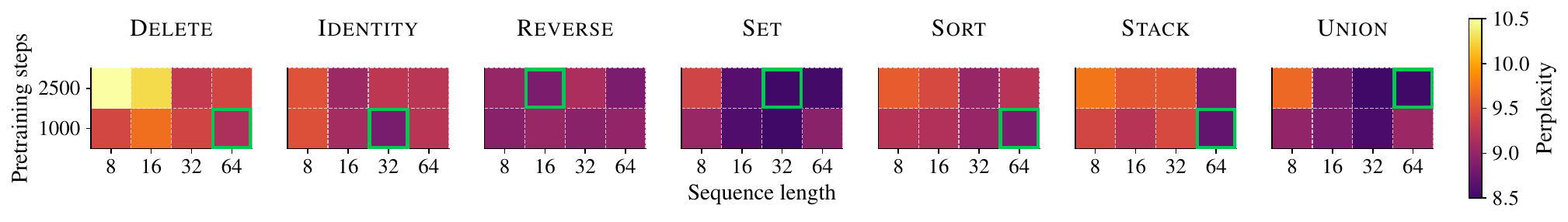}
    \caption{
Validation perplexity for different configurations of procedural pretraining when finetuned on \textsc{WikiText} (top) and \textsc{JavaCorpus} (bottom), 
sweeping over sequence length and number of pretraining steps. 
Each panel corresponds to a distinct procedural task, with colours indicating perplexity 
(lower is better). The best-performing configuration for each task is marked in green. 
}
    \label{fig:grid_search_wiki_java}
\end{figure}

\begin{figure}[H]
    \centering
    \includegraphics[width=0.37\linewidth]{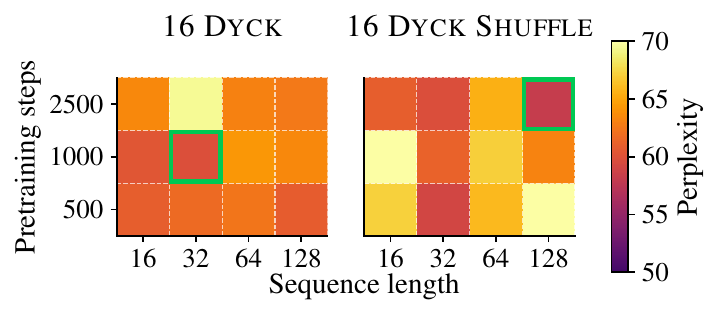}
    \qquad
    \includegraphics[width=0.37\linewidth]{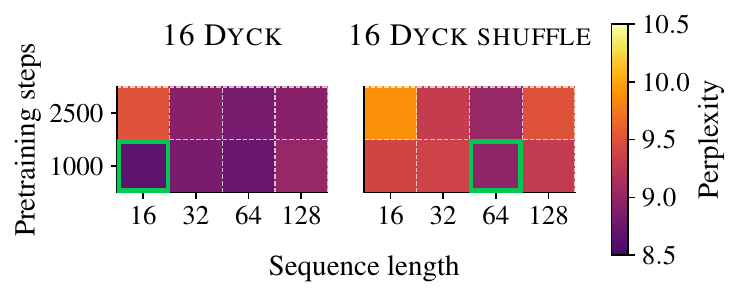}
    \caption{Validation perplexity for \textsc{Dyck} and \textsc{Dyck Shuffle} procedural pretraining when fine-tuned on \textsc{WikiText} (left) and \textsc{JavaCorpus} (right), sweeping over sequence length and number of pretraining steps. Setup matches Figure~\ref{fig:grid_search_wiki_java}. Colours indicate perplexity (lower is better), with the best-performing configuration marked in green.}
    \label{fig:grid_search_wiki_java_dyck}
\end{figure}

\paragraph{Results.} 
Figure~\ref{fig:grid_search_wiki_java} and \ref{fig:grid_search_wiki_java_dyck} report validation perplexity across all configurations, showing that both sequence length and pretraining steps strongly influence performance, with optimal settings differing by domain and task.

% \section{\red{Learning Rate Matching}}\label{app:learning_rate_matching}
% \begin{itemize}
%     \item 
% \end{itemize}

\section{Longer Sequences for Procedural Pretraining}
\label{app:longer_sequences}

We extend the sequence length search on \textsc{WikiText} from 8–64 tokens (Appendix~\ref{app:grid-search}) to 128 tokens using full-model transfer for the best perfoming procedural tasks. Results are mixed: \textsc{Set} benefits from longer sequences, while \textsc{Sort} and \textsc{Union} do not. Thus, the utility of longer procedural sequences is task-dependent.

\begin{figure}[h!]
    \centering
    \includegraphics[width=0.55\linewidth]{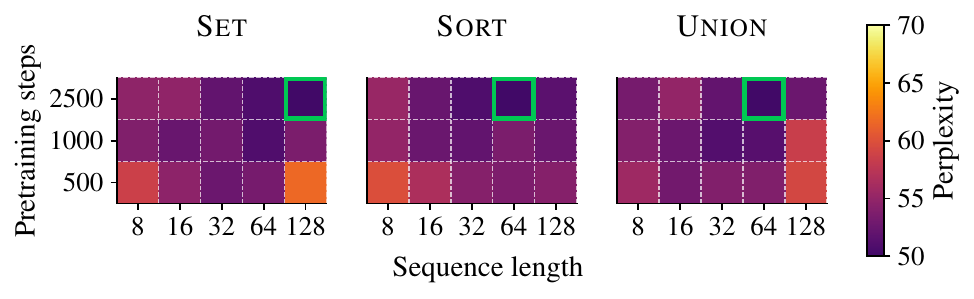}
    \caption{Effect of extending sequence length during procedural pretraining on \textsc{WikiText}. 
    Longer sequences improve subsequent language modelling for \textsc{Set} but not \textsc{Sort} or \textsc{Union}, showing that the benefit of extended contexts is task-dependent.}

\end{figure}

\section{Transferability Analysis}

We analyse the correlation between procedural pretraining loss and downstream  loss on \textsc{C4}.
For \textsc{Set} and \textsc{Union}, transfer performance deteriorates when procedural loss is either too high or too low, suggesting that both underfitting and overfitting impair generalization. Consequently, the strongest transfer is observed at intermediate levels of procedural optimization. In contrast, for \textsc{Sort}, transfer performance contintues to improve steadily as procedural loss decreases, demonstrating that the transferability of procedural pretraining is task dependent.

\label{app:transferability_analysis}
\begin{figure}[h!]
    \centering
    \includegraphics[width=0.3\linewidth]{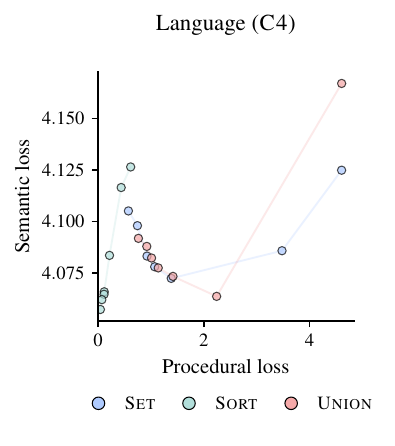}
    \caption{Transferability of procedural pretraining. 
    Relationship between procedural validation loss and downstream loss on \textsc{C4}. 
    For \textsc{Set} and \textsc{Union}, transfer is strongest at intermediate procedural losses, with both underfitting and overfitting harming generalization. 
    For \textsc{Sort}, continually decreasing procedural loss consistently improves transfer.}

    \label{fig:transferability-loss-to-loss}
\end{figure}

\section{The Effect of Vocabulary Size}
We investigate the effect of vocabulary size during procedural pretraining. 

\paragraph{Setup.} 
Models are pretrained on \textsc{Set}, \textsc{Sort}, and \textsc{Union} with vocabularies from 25 to 500 symbols (the main results use 100 by default), then transferred to \textsc{WikiText} using full-model transfer. Evaluation perplexity is reported after fine-tuning.

\label{app:vocab_size}
\begin{figure}[h!]
    \centering
    \includegraphics[width=0.4\linewidth]{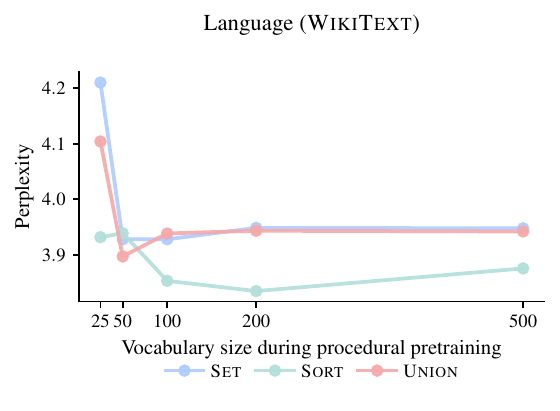}
        \caption{Effect of vocabulary size during procedural pretraining on \textsc{WikiText}. 
    Small vocabularies (25–50) degrade transfer performance, while moderate sizes ($\sim$100-200) are sufficient. 
    Larger vocabularies offer no further improvement.}
    \label{fig:vocab_ablation}
\end{figure}

\paragraph{Results.} 
As shown in Figure~\ref{fig:vocab_ablation}, very small vocabularies (25–50) harm transfer, leading to higher perplexity. 
For \textsc{Set} and \textsc{Union}, performance stabilizes once the vocabulary reaches a moderate size ($\sim$100), with larger sizes offering no further gains.
\textsc{Sort} benefits modestly at 200 but declines at 500. 
Overall, procedural pretraining is most effective within a moderate vocabulary range, too small harms transfer, while too large brings no improvement or negative return.

% % \section{\red{The Effect of Warmup}}
% % \label{app:warmup}
% % To Do.

\section{Weight Decay Ablation}
\label{app:weight-decay}
In the main paper, natural language experiments use a weight decay of 0.1 during procedural pretraining, following \citet{hu2025between}. 
To test this choice, we reduce the weight decay to 0.01 (the value used for code and math) and evaluate performance on \textsc{C4} semantic pretraining. 
The takeaway that MLP-only transfer is best for natural language remains unchanged, showing that our findings are robust to this hyperparameter.

\label{app:weight_decay}
\begin{figure}[h!]
    \centering
    \includegraphics[width=0.5\linewidth]{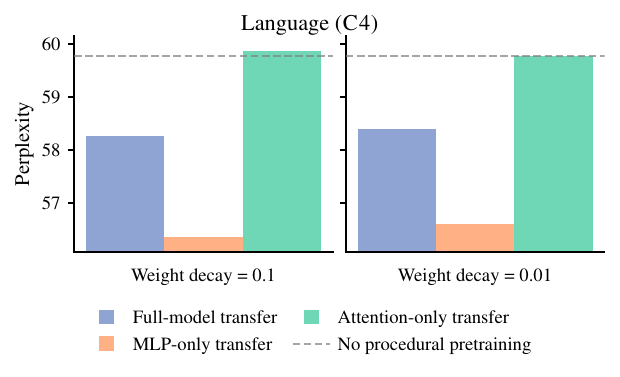}
    \caption{Effect of weight decay during procedural pretraining on \textsc{C4}. 
Changing weight decay from 0.1 to 0.01 does not alter the outcome: MLP-only transfer remains the best configuration for natural language.}
    \label{fig:weight-decay-ablation}
\end{figure}

% \section{Additional Results on CodeSearchNet, WebText, DeepMind-Math}
% \label{app:additional_results}
% To Do.

% \input{9_related_work_long}

\section{Scaling Procedural Pretraining}
\label{app:scaling-procedural-pretraining}
\reb{Extending the findings of Section~\ref{sec:scaling_up}, we scale both model size and semantic pretraining data size.  We increase the architecture to 350M parameters, and further to a 1.3B-parameter model (architectural hyperparameters follow~\citet{biderman2023pythia}), while scaling natural-language pretraining to 1.6B / 6.6B \textsc{C4} tokens and 4.8B / 10.5B \textsc{CodeParrot} tokens respectively.} 

\reb{ 
For 350M models and 1.3B models, we use a learning rate of $3\times10^{-4}$ and $2\times10^{-4}$, following~\citet{biderman2023pythia}.
We also use larger batch sizes and/or training steps for the semantic pretraining to increase the semantic tokens.
Other hyperparameters follow Appendix~\ref{app:experimental-details}.
We utilise \textsc{Union} for procedural pretraining on both \textsc{C4} and \textsc{CodeParrot}.}

\reb{\textbf{Additive setting.}
We find procedurally pretrained models continue to substantially outperform their non-procedural counterparts across all scales (Table~\ref{tab:gptmedium-c4-code}). This shows that the benefits of procedural pretraining persist at substantially larger scales in both model capacity and dataset size.
}

\begin{table}[H]
\centering
\small
\begin{tabular}{lcc}
\toprule
\reb{Model} & \reb{\textsc{C4} (Perplexity $\downarrow$)} & \reb{\textsc{CodeParrot} (Perplexity $\downarrow$)} \\
\midrule

\multicolumn{3}{c}{\textbf{350M parameters}} \\
\midrule
\reb{\textsc{No procedural pretraining}} & \reb{40.3} & \reb{4.97} \\
\reb{\textsc{Ours (Union)}}              & \reb{\textbf{39.0}} & \reb{\textbf{4.62}} \\

\midrule
\multicolumn{3}{c}{\textbf{1.3B parameters}} \\
\midrule
\reb{\textsc{No procedural pretraining}} & \reb{28.8} & \reb{3.45} \\
\reb{\textsc{Ours (Union)}}              & \reb{\textbf{27.3}} & \reb{\textbf{3.36}} \\

\bottomrule
\end{tabular}
\vspace{1em}
\caption{\reb{
Perplexity of language models with and without procedural pretraining at increased scale.
350M-parameter models are pretrained on 1.6B \textsc{C4} tokens and 4.8B \textsc{CodeParrot} tokens.
1.3B-parameter models are pretrained on 6.6B \textsc{C4} tokens and 10.5B \textsc{CodeParrot} tokens.
\textbf{Procedural pretraining consistently improves perplexity across both scale regimes.}
}}
\label{tab:gptmedium-c4-code}
\end{table}

\reb{We additionally report BLiMP evaluation for the larger C4-trained models. These show that procedural pretraining imparts  lasting gains in syntactic and morphological generalization at a larger scale (Table~\ref{tab:blimp_procedural_vs_baseline}).}

\begin{table}[H]
\centering
\small
\begin{tabular}{lc}
\toprule
\reb{Model} & \reb{BLiMP (Accuracy $\uparrow$)} \\
\midrule

\multicolumn{2}{c}{\textbf{350M parameters}} \\
\midrule
\reb{\textsc{No procedural pretraining}} & \reb{71.5} \\
\reb{\textsc{Ours (Union)}}              & \reb{\textbf{72.9}} \\

\midrule
\multicolumn{2}{c}{\textbf{1.3B parameters}} \\
\midrule
\reb{\textsc{No procedural pretraining}} & \reb{73.2} \\
\reb{\textsc{Ours (Union)}}              & \reb{\textbf{75.5}} \\

\bottomrule
\end{tabular}
\vspace{1em}
\caption{\reb{
BLiMP accuracy for language models with and without procedural pretraining at increased scale. \textbf{Procedural pretraining consistently improves grammatical acceptability across both scales.}
}}
\label{tab:blimp_procedural_vs_baseline}
\end{table}

\reb{\textbf{Substitutive setting.} We further evaluate the substitutive setting at the 1.3B-parameter scale. 
Specifically, we use only 82M procedural tokens. 
Despite this minimal additional data, procedural pretraining enables the model to match baseline performance using just 66\% of the \textsc{C4} data and 75\% of the \textsc{CodeParrot} data. 
This corresponds to a reduction of 2.1B \textsc{C4} tokens and 2.5B \textsc{CodeParrot} tokens in semantic pretraining.}

\section{Results on downstream tasks}
\label{app:downstream_finetuning}

\reb{This section provides the extended downstream fine-tuning or few-shot results referenced in Section~\ref{sec:scaling_up}. See Appendix~\ref{app:downstream_setup} for additional experimental details.}

\subsection{Finetuning Results}
\reb{\textbf{Setup.} 
To investigate whether the benefits of procedural pretraining persist after downstream fine-tuning, we conduct an additional fine-tuning step. Specifically, we finetune the language models (pretrained on \textsc{C4}) on both  \textsc{Wikitext-103} and \textsc{GLUE} tasks independently. The code models (pretrained on \textsc{CodeParrot}) are finetuned and evaluated on \textsc{PY150}. For WikiText-103, we use the \textsc{Sort} model, as it obtains the lowest perplexity on \textsc{C4}. 
For GLUE and PY150, we instead use the \textsc{Union} model as it has demonstrated consistently strong performance across a broad range 
of downstream tasks. }

\reb{
\textbf{Results.} Consistent with the main findings of enhancing semantic pretraining, the procedurally pretrained models continue to outperform the baseline across these downstream tasks (Table~\ref{tab:finetuning-wiki103-py150} and Table~\ref{tab:glue_procedural}). This shows that the benefits of procedural data persist after fine-tuning on downstream tasks, suggesting the potential of using procedural pretraining to enhance the practical utility of models.
}

\begin{table}[H]
\centering
\small
\begin{tabular}{lcc}
\toprule
\reb{Model} & \reb{\textsc{WikiText-103} (Perplexity $\downarrow$)} & \reb{\textsc{PY150} (Accuracy $\uparrow$)} \\
\midrule
\reb{\textsc{No Procedural pretraining}} & \reb{33.0} & \reb{60.5} \\
\reb{\textsc{Ours}} & \reb{\textbf{32.3}} & \reb{\textbf{62.1}} \\
\bottomrule
\end{tabular}
\vspace{1em}
\caption{\reb{Downstream fine-tuning results on \textsc{WikiText-103} (perplexity; after \textsc{C4} pretraining) and \textsc{PY150} (accuracy; after \textsc{CodeParrot} pretraining), comparing models with and without procedural pretraining.}}
\label{tab:finetuning-wiki103-py150}
\end{table}

\begin{table}[H]
\centering
\small
\begin{tabular}{lccccccccc|c}
\toprule
& \reb{COLA} & \reb{SST-2} & \reb{MRPC} & \reb{QQP} & \reb{STS-B} & \reb{MNLI} & \reb{QNLI} & \reb{RTE} & \reb{WNLI} & \textbf{\reb{Avg}} \\
\midrule
\textsc{\reb{No Proc. P.T.}} 
& \reb{69.1} & \reb{85.3} & \reb{70.8} & \reb{84.3} & \reb{55.1} & \reb{72.1} & \reb{79.9} & \reb{57.4} & \reb{42.3} & \reb{68.5} \\
\reb{\textsc{Ours}}
& \reb{68.9} & \reb{87.6} & \reb{69.6} & \reb{84.8} & \reb{68.8} & \reb{72.7} & \reb{81.3} & \reb{55.6} & \reb{52.1} & \textbf{\reb{71.3}} \\
\bottomrule
\end{tabular}
\vspace{1em}
\caption{\reb{GLUE scores after C4 pretraining, comparing the baseline without procedural pretraining to our model with procedural pretraining.}}
\label{tab:glue_procedural}
\end{table}

\subsection{Zero-shot Results}

\paragraph{Setup.}
We further evaluate whether the benefits of procedural pretraining persist in zero-shot downstream evaluation. We compare
models with and without procedural pretraining on commonsense reasoning benchmarks, using ARC-Easy~\citep{clark2018think}, HellaSwag~\citep{zellers2019hellaswag}.

\paragraph{Results.}
Consistent with the fine-tuning results in Appendix~M.1, procedural pretraining improves performance on both commonsense
reasoning benchmarks (Table~\ref{tab:commonsense_results}). These results indicate that the gains from procedural data
also transfer to zero-shot commonsense reasoning.

\begin{table}[t]
\centering
\caption{Zero-shot commonsense reasoning results on ARC-Easy and HellaSwag, comparing models with and without procedural pretraining.}
\label{tab:commonsense_results}
\begin{tabular}{lcc}
\toprule
Model & ARC-Easy (Accuracy $\uparrow$) & HellaSwag (Accuracy $\uparrow$) \\
\midrule
\textsc{No Procedural Pretraining} & 37.3 & 27.6 \\
\textsc{Ours} & \textbf{38.2} & \textbf{28.3} \\
\bottomrule
\end{tabular}
\end{table}

\section{Additional Results}
\label{app:more_results}
\input{9_more_results}

%% file: 9_related_work_long.tex
\section{Extended Literature Review}
\label{app:related_work}

% Copy from workshop paper below.

%-------------------------------------------------------
% \todo{Z: I added this back, currently a copy from workshop paper}
\paragraph{What is learned by pretraining language models.}
The quantity~\citep{kaplan2020scaling} and quality
\citep{longpre2024pretrainer}
%\citep{goyal2024scaling}
of pretraining data are empirically critical for the performance of large language models.
But recent results also question the value of the data,
%used for pretraining
%vs.\ the optimization procedure and objective themselves.
showing that some benefits of pretraining are attributable to the optimisation objective more than the actual data.
\citet{balestriero2024perception} % For Perception Tasks: Is LLM Pretraining by Next-Token Prediction Worth its Cost?
compared models trained for text classification from random initialisation
with fine-tuning from a pretrained checkpoint. They found that pretraining provides little benefit for tasks that do not involve text generation.
\citet{krishna2023downstream} % Downstream Datasets Make Surprisingly Good Pretraining Corpora
showed success in re-using the same data for pretraining and fine-tuning,
showing also that the pretraining objective matters more than the data being used.
The same conclusion follows from results of pretraining on synthetic data devoid of semantic meaning,
e.g.\ for machine translation~\citep{he2023synthetic}, % Synthetic Pretraining Tasks for Neural Machine Translation (simple toy synthetic data for translation model: gets some of the pretraining benefits without any semantic meaning),
computer vision~\citep{baradad2021learning}, % Learning to See by Looking at Noise
visual navigation~\citep{wang2022visual}, % Visual Pretraining for Navigation: What Can We Learn from Noise?
and reinforcement learning~\citep{baradad2022procedural}. % Procedural Image Programs for Representation Learning
This paper examines such purely synthetic pretraining to understand the exact capabilities that can be obtained from procedurally-generated data.
%pushes this idea of  further and characterises the kind of information gained from various forms of synthetic pretraining data.

%-------------------------------------------------------
\paragraph{What matters in pretraining data.}
The selection of data to pretrain frontier models
mostly relies on experimentation%
%and broad guidelines, e.g.\ for a diversity of sources and domains
~\citep{longpre2024pretrainer}.
However, several key distributional and structural properties of the data
have also been identified, such as data repetition to foster generalisation~\citep{charton2024emergent}
and burstiness to enable in-context learning~\citep{chan2022data}.
Computer code is empirically very effective as pretraining data for LLMs,
as it improves their abilities for compositional generalisation and
math-related tasks~\citep{aryabumi2024code,petty2024does}.
% To code, or not to code? exploring impact of code in pretraining
% How Does Code Pretraining Affect Language Model Task Performance?
This presumably results from the abundant compositional and recursive patterns in computer code,
but a better understanding of the mechanisms at play
is lacking to reap the full benefits of structure in pretraining data.
%The empirical evidence is compelling in LLMs at the largest scale.
%these effects are difficult to study in frontier-scale models.
In this paper, we replicate the positive effects of structured pretraining data in controlled settings,
and study how such data imparts useful inductive biases to the model.

%These results add to recent demonstrations using procedural data~\citep{hu2025between}
 % Between circuits and chomsky: Pre-pretraining on formal languages imparts linguistic biases
%formal lang data more valuable than natural lang for pretraining.
%to suggest that some value of pretraining depends on structural properties of the data more than its actual semantic contents.

%-------------------------------------------------------
\paragraph{Pretraining on procedural data.}
Most attempts to train language models with synthetic data
follow a linguistic perspective,
using formal languages to imitate properties of natural language
\citep{chiang2022transferability,goodale2025meta,mccoy2023modeling,papadimitriou2023injecting,ri2022pretraining}.
% Pretraining with artificial language: Studying transferable knowledge in language models (encoder models; learns using context from formal lang)
% On the transferability of pretrained language models: A study from artificial datasets (token pairs, GLUE RoBERTa encoders)
Recent work considers increasingly simpler forms of synthetic data such as input/outputs of simple algorithms
\citep{lindemann2024sip,wu2022insights}.
In these papers, specific forms of synthetic pretraining data prove helpful to subsequent fine-tuning on natural language tasks. 
\citet{hu2025between} provide strong empirical benefits,
% Between circuits and chomsky: Pre-pretraining on formal languages imparts linguistic biases
showing that data generated from formal languages is more valuable token-per-token than natural language
for training a 1B-parameter language model.
\citet{zhang2024intelligence} pretrain on traces of cellular automata and show marginal but consistent improvements on simple reasoning tasks.
% Intelligence at the Edge of Chaos (trained on synthetic data, embeddings ft on other tasks; not learning the simplest generalizing rule)
Our study complements this line of work by examining more closely the pretrained models
%on procedural data
on diagnostic tasks, rather than evaluating their general handling of natural language.
We identify specific capabilities imparted by specific types of procedural tasks,
and locate useful structure in different parts of the architecture.
We also investigate methods to combine the benefits from multiple complementary tasks.

%\citep{papadimitriou2023injecting} % Injecting structural hints: Using language models to study inductive biases in language learning (test hypotheses for human inductive bias by pretr LMs on formal lang)
%best: recursive processing + a bit of context sensitivity + Zipfian vocabulary distribution

%\citep{lindemann2024sip}
% SIP: Injecting a Structural Inductive Bias into a Seq2Seq Model by Simulation (train transformer to execute programs; specific to transduction (sequence-to-sequence) tasks, encoder-decoder transformers)

%\citep{wu2022insights}
% Insights into pretraining via simpler synthetic tasks

%\citep{mccoy2023modeling}
% Modeling rapid language learning by distilling Bayesian priors into artificial neural networks
%and follow-up:
%\citep{goodale2025meta}
% Meta-Learning Neural Mechanisms rather than Bayesian Priors
%sort of pretraining (meta learning) formal languages to acquire mechanisms (e.g.\ a counting ability) that is useful (as a sort of inductive bias) for subsequent learning of language

%This paper explores the limits of the benefits that can be gained by pretraining on such synthetic data.

\paragraph{Procedural data in vision and RL.}
%In \textbf{computer vision}, 
Vision transformers (ViTs)
have been trained on synthetic data
of increasingly simple nature~\citep{baradad2021learning}.
\citet{nakamura2024scaling} % Scaling Backwards: Minimal Synthetic Pretraining?
pretrained ViTs on a single fractal image with augmentations
that remarkably match the performance of ImageNet-pretrained models after fine-tuning.
This indicates that %the pretraining procedure and
structural properties of the data
matter more than its semantic contents.
Similar results exist in {reinforcement learning} with models pretrained
on data generated from random Markov chains
\citep{wang2023pretraining} % Pretraining with Synthetic Data Helps Offline Reinforcement Learning
and noise-based images
\citep{baradad2022procedural}. % Procedural Image Programs for Representation Learning

%-------------------------------------------------------
%\paragraph{Transferring parts of pretrained models.}
\paragraph{Partial transfer from pretrained transformers.}
%Rather than using a complete pretrained model for fine-tuning,
\citet{zhang2023instilling} % Instilling inductive biases with subnetworks
and
\citep{xu2023initializing}
% Initializing Models with Larger Ones
showed that copying
subsets of pretrained weights
%specific subnetworks from a pretrained model
could transfer specific capabilities. %to a downstream model.
\citet{abnar2020transferring} % Transferring Inductive Biases through Knowledge Distillation
used knowledge distillation to transfer the inductive biases of one architecture into another.
The ``mimetic initialisation'' of self-attention
\citep{trockman2023mimetic}
is a procedure  handcrafted to imitate the locality bias of pretrained models.
We also evaluate the partial transfer of pretrained weights,
which reveals that different pretraining tasks create useful structure in different parts of the architecture.

% Mimetic Initialization of Self-Attention Layers (novel init for ViTs; mimics pretrained ones, very basic)
%\citep{d2021convit}
% ConViT: Improving Vision Transformers with Soft Convolutional Inductive Biases (init self-attention to mimic convolutiions)
%\citep{zheng2024structured}
% Structured Initialization for Attention in Vision Transformers (KQ initialized to for att to mimic one-hot conv matrices)
%\citep{trockman2022understanding}
% Understanding the Covariance Structure of Convolutional Filters (init of conv filters in ConvMixer and ConvNeXt)

%-------------------------------------------------------
\paragraph{Pretraining as an inductive bias.}
Pretraining transformers on synthetic data has been used to
mimic the inductive biases of Bayesian inference %Gaussian processes
\citep{muller2021transformers} % Transformers can do bayesian inference (Prior-Fitted Networks; meta learn few-shot prediction in one forward pass)
%Prior-Data Fitted Networks (PFNs)
and Solomonoff Induction
\citep{grau2024learning}. % Learning universal predictors (meta learning from synthetic data from Turing machines)
%\citep{dorrell2023meta}
% Meta-learning the inductive bias of simple neural circuits
\citet{goodale2025meta} showed that well-chosen formal languages can teach complex mechanisms (e.g.\ counters) to a sequence model.
Pretraining can generally be seen as a \emph{soft} inductive bias for subsequent fine-tuning.
But there is a large gap in our understanding of its effects compared to those of
\emph{hard} inductive biases
of neural architectures~\citep{teney2024neural,teney2025we}.
%Finally,
\citet{han2025position} argue that the difficulties of LLMs to reason robustly
is due to their entangled representation of knowledge and reasoning.
Much remains to be understood about how both are learned from the same data~\citep{ruis2024procedural}.
Our results suggest that procedural data could be one way to acquire
reasoning mechanisms independently from specific pieces of knowledge.

%% file: 9_procedural_attention_analysis.tex
This appendix analyses whether the benefits of procedural pretraining arise from generic attention sharpening.
First, we find that a small subset of sharpened attention heads contain the useful inductive bias for downstream tasks. Then, we attempt to reproduce the behaviour of these heads through regularisation. We find this does not provide the same benefits, demonstrating that procedural pretraining fosters specific inductive biases beyond generic attention sharpening.

\begin{figure}[H]
    \centering
    \includegraphics[width=1.0\linewidth]{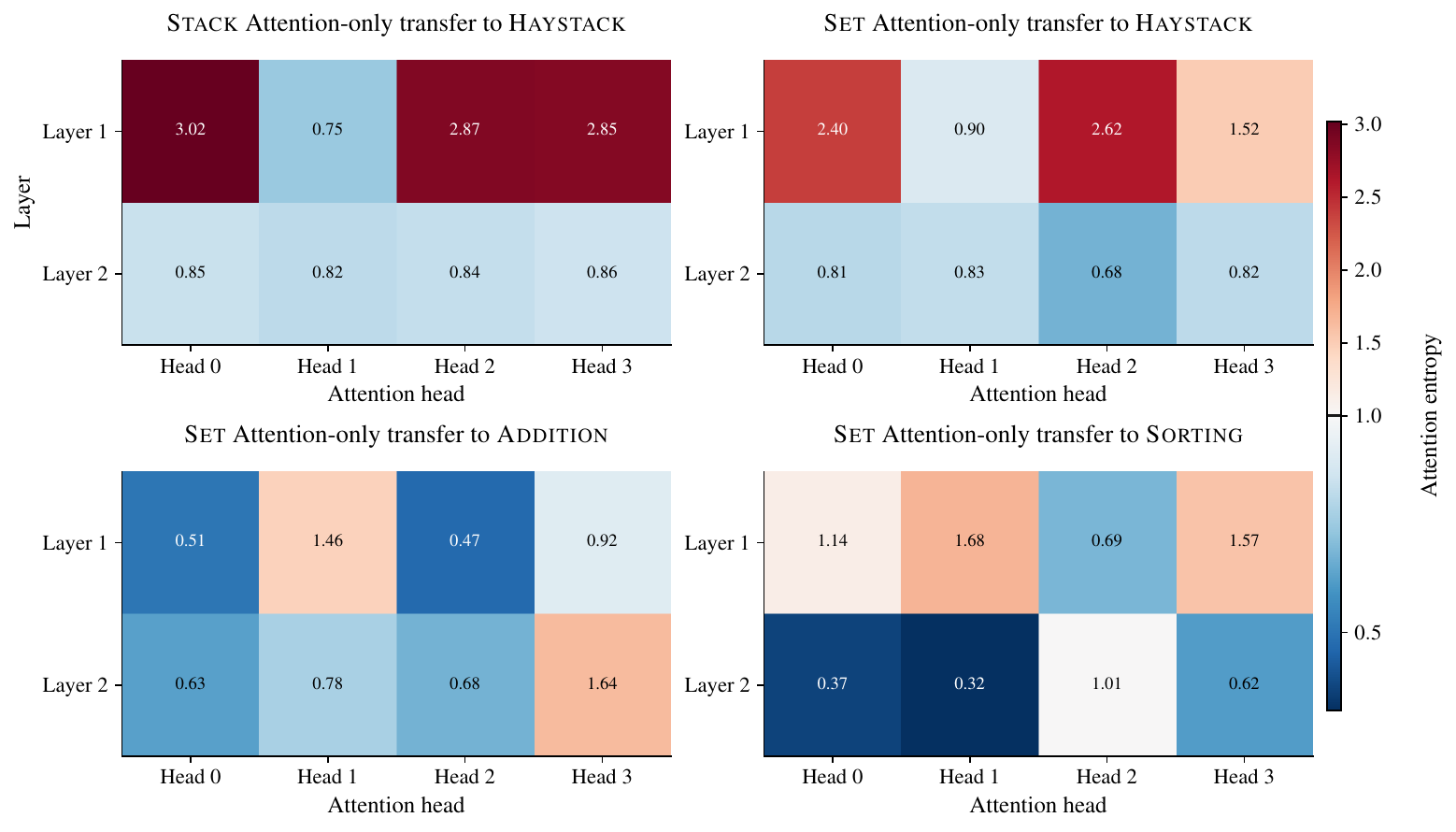}
    \caption{Head-wise attention entropy after fine-tuning. Procedural pretraining yields a subset of low-entropy heads (blue).}
    \label{fig:procedural_attention_analysis_heatmap}
\end{figure}

\subsubsection{Attention Entropy Analysis}
\label{app:procedural-attention-analysis_entropy-heatmap}
We first examine the attention patterns of the procedurally pretrained models after fine-tuning on the downstream tasks.

\textbf{Setup.}
We measure the sharpness of each attention head using entropy,  
\[
H = -\sum_{i} p_i \log p_i,
\]
where \(p_i\) denotes the normalized attention weight assigned to token \(i\). 
Low entropy corresponds to selective attention, while high entropy reflects diffuse, uniform distributions. 
We compute head-wise entropy after fine-tuning, averaging over 100 downstream evaluation examples.

\textbf{Results.}
Figure~\ref{fig:procedural_attention_analysis_heatmap} shows that procedural pretraining leads models, after downstream fine-tuning, to consistently develop a subset of low-entropy heads.
For example, a \textsc{Stack}-pretrained model fine-tuned on \textsc{Haystack} exhibits five of eight heads with entropy close to \(H \approx 0.8\), while the remaining three have substantially higher entropy around \(H \approx 3.0\).

\subsubsection{Selective Transfer of Low-Entropy Heads}
\label{app:entropy-head-transfer}
We hypothesise that the useful inductive biases introduced by procedural pretraining are concentrated in the subset of low-entropy attention heads.

\textbf{Setup.}
To test our hypothesis, we fine-tune on the downstream task while transferring either the three lowest-entropy heads that emerge from the procedurally pretrained model (identified post hoc after finetuning) or, for comparison, the three highest-entropy heads.

\begin{figure}[H]
    \centering
    \includegraphics[width=0.7\linewidth]{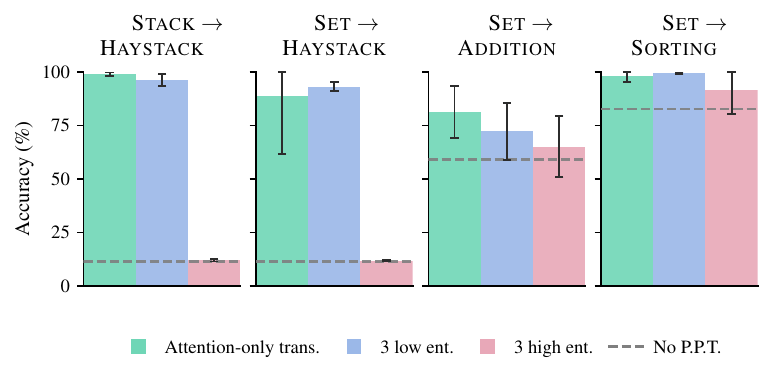}
    \caption{Validation accuracy after downstream fine-tuning when transferring subsets of procedurally pretrained attention heads. The three lowest-entropy heads preserve or even surpass full transfer, while the three highest-entropy heads perform comparably to a baseline without procedural pretraining. Results are over 10 random seeds.}
    \label{fig:procedural_attention_analysis_head_transfers}
\end{figure}

\paragraph{Results.} 
Figure~\ref{fig:procedural_attention_analysis_head_transfers} shows that transferring only the three lowest-entropy heads preserves, and in some cases even surpases the performance of full attention transfer. 
In contrast, transferring the three highest-entropy heads results in performance comparable to the baseline without procedurally pretrained attention.
These results demonstrate that the benefits of procedural pretraining can be concentrated in a small subset of sharp, low-entropy attention heads.

\subsubsection{Entropy Regularisation to Selected Attention Heads}
\label{app:head-entropy-regularisation}

We next investigate whether the benefits of procedural pretraining can be reproduced by explicitly enforcing low-entropy attention.

\textbf{Setup.}
We attempt to replicate the behavior of the beneficial attention heads through regularization. An entropy regularization term is introduced during finetuning on \textsc{Haystack} to a model that did not undergo procedural pretraining. This regularization is applied to three selected heads and drives them toward a target entropy of $\tau = 0.8$, matching the average entropy observed in the three heads shown to carry useful inductive biases from \textsc{Stack} pretraining (Figure~\ref{fig:procedural_attention_analysis_heatmap} and Figure~\ref{fig:procedural_attention_analysis_head_transfers}).

\begin{figure}[t]
    \centering
    \begin{minipage}[t]{0.35\linewidth}
        \vspace{0pt}
        \centering
        \includegraphics[width=\linewidth]{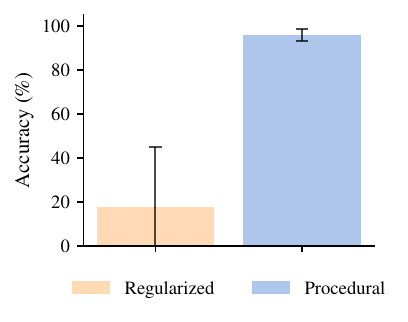}
    \end{minipage}\hfill
    \begin{minipage}[t]{0.6\linewidth}
        \vspace{0pt}
        \caption{\textbf{Validation accuracy on \textsc{Haystack} with entropy regularisation.}
        Models trained from scratch with explicitly enforced low-entropy heads (orange) underperform those with procedurally pretrained heads (blue), indicating that sharper attention alone is insufficient. Results are averaged over 10 random seeds.}
        \label{fig:procedural_attention_analysis_head_regularisation}
    \end{minipage}
\end{figure}

\textbf{Results.}
As shown in Figure~\ref{fig:procedural_attention_analysis_head_regularisation}, this approach is ineffective: the regularized heads perform substantially worse than the \textsc{Stack}-pretrained heads when evaluated on the \textsc{Haystack} task.

In summary, these findings indicate that simply enforcing sharper attention is insufficient to reproduce the benefits of procedural pretraining. Low entropy appears to be a side effect of the inductive biases acquired through procedural pretraining rather than the cause of improved performance.

%% file: 9_more_results.tex
\subsection{Algorithmic Reasoning Tasks}
\label{app:more_results-algorthmic-reasoning}
\begin{table}[H]
\vskip 0.15in
\begin{center}
\begin{small}
\begin{sc}
\begin{tabular}{lcccccc}
\toprule
\textnormal{Pretraining task} & Haystack & Addition & Reversed addition & Multiplication & Sorting  \\
\midrule
Rand init.                & $11.3 \pm 0.4$ & $59.1 \pm 7.0$ & $76.4 \pm 23.2$ & $42.7 \pm 5.3$ & $82.7 \pm 11.6$\\
\midrule
4-Dyck              & $98.3 \pm 1.1$ & $52.7 \pm 0.3$ & $35.7 \pm 2.5$  & $46.7 \pm 4.6$ & $56.3 \pm 19.2$\\
8-Dyck              & $93.6 \pm 1.3$ & $53.4 \pm 0.3$ & $48.9 \pm 4.9$  & $44.5 \pm 0.9$ & $98.7 \pm 0.3$ \\
16-Dyck             & $96.9 \pm 1.0$ & $87.8 \pm 4.2$ & $83.5 \pm 0.6$  & $39.4 \pm 3.3$ & $95.5 \pm 1.0$ \\
\midrule
4-Dyck shuffle       & $7.3 \pm 0.6$  & $54.5 \pm 0.2$ & $87.8 \pm 12.9$ & $41.8 \pm 3.7$ & $61.0 \pm 1.4$ \\
8-Dyck shuffle       & $9.6 \pm 0.3$  & $67.7 \pm 0.8$ & $90.1 \pm 5.9$  & $37.4 \pm 0.1$ & $84.1 \pm 5.7$ \\
16-Dyck shuffle      & $18.6 \pm 26.3$& $70.8 \pm 5.5$ & $87.0 \pm 12.8$ & $44.0 \pm 0.1$ & $71.1 \pm 5.4$\\
\midrule
Stack               & $55.2 \pm 39.3$ & $62.3 \pm 5.3$ & $34.9 \pm 0.2$  & $46.6 \pm 2.0$ & $21.3 \pm 0.6$  \\
\midrule
Identity            & $18.8 \pm 14.3$ & $54.7 \pm 0.2$ & $42.7 \pm 0.9$  & $46.6 \pm 2.7$ & $19.9 \pm 0.5$  \\
\midrule
Set                 & $18.9 \pm 26.6$ & $53.4 \pm 0.1$ & $44.6 \pm 5.1$  & $43.5 \pm 8.4$ & $93.5 \pm 1.6$  \\
\midrule
Union & $9.8 \pm 1.1$  & $48.6 \pm 0.7$  & $50.8 \pm 0.2$ & $63.5 \pm 2.3$  & $16.9 \pm 0.5$  \\
\midrule
Reverse & $33.3 \pm 22.4$  & $46.1 \pm 2.3$ & $46.8 \pm 1.33$ & $54.4 \pm 3.2$ & $16.7 \pm 0.5$  \\
\midrule
Delete  & $52.6 \pm 22.4$  & $60.7 \pm 4.19$  & $40.0 \pm 1.8$ & $61.9 \pm 1.4$ & $20.1 \pm 0.6$ \\

\midrule
\textsc{Eca rule 110}                & $10.5 \pm 0.5$ & $69.6 \pm 7.9$ & $91.1 \pm 16.1$ & ---            & $76.9 \pm 1.4$  \\
\midrule
Best model shuffled & $10.3 \pm 0.5$ & $52.0 \pm 0.3$ & $65.0 \pm 21.4$& $48.4 \pm 4.4$ & $69.9 \pm 2.2 $\\

\bottomrule
\end{tabular}
\end{sc}
\end{small}
\end{center}
\vspace{1em}
\caption{Full results across all pretraining tasks and algorithmic reasoning tasks. Each cell reports the mean accuracy $\pm$ standard deviation over 10 random seeds, except for \textsc{Multiplication}, which is over 3 seeds. The means of these results are visualised in Figure~\ref{fig:alg_reasoning_results}.}

\label{tab:full_res}
\end{table}

%Needle In A Haystack Component Transfers 
\begin{table}[H]
\label{tab:NIAH-component-transfer}
\vskip 0.15in
\begin{center}
\begin{small}
\begin{sc}
\begin{tabular}{lccc}
\toprule
\textnormal{Pretraining task} & Full transfer & MLP only & Attention only \\
\midrule
4-Dyck            & $98.3 \pm 1.1$   & $8.7 \pm 0.5$   & $11.6 \pm 0.5$ \\
16-Dyck shuffle    & $18.6 \pm 26.3$  & $8.9 \pm 0.9$   & $16.5 \pm 10.6$ \\
Stack             & $55.2 \pm 39.3$  & $7.1 \pm 0.6$   & $98.9 \pm 0.8$ \\
Identity          & $18.8 \pm 14.3$  & $7.0 \pm 0.9$   & $99.0 \pm 1.7$ \\
Set               & $18.9 \pm 26.6$  & $8.3 \pm 0.7$   & $88.9 \pm 27.1$ \\
Union      & $9.8 \pm 1.1$  & $8.2 \pm 0.7$ & $11.7 \pm 0.4$  \\
Reverse & $33.3 \pm 22.4$  & $7.3 \pm 1.2$ & $98.6 \pm 0.8$  \\
Delete & $52.6 \pm 22.4$  & $8.4 \pm 0.8$ & $91.8 \pm 3.5$  \\
ECA               & $10.5 \pm 0.5$   & $8.7 \pm 1.0$   & $11.6 \pm 1.0$ \\
\bottomrule
\end{tabular}
\end{sc}
\end{small}
\end{center}
\vspace{1em}
\caption{
\textsc{Haystack} task accuracy (mean $\pm$ standard deviation over 10 seeds) for models initialized with weights from different pretraining tasks. We report results for full model transfer, MLP-transfer, and attention-transfer.
}
\end{table}

% Addition Component Transfers
\begin{table}[H]
\label{tab:addition-component-transfer}
\vskip 0.15in
\begin{center}
\begin{small}
\begin{sc}
\begin{tabular}{lccc}
\toprule
\textnormal{Pretraining task} & Full transfer & MLP only & Attention only \\\midrule
16-Dyck           & $87.8 \pm 4.2$   & $60.0 \pm 6.6$   & $59.2 \pm 10.4$ \\
16-Dyck shuffle    & $70.8 \pm 5.5$   & $61.7 \pm 6.9$   & $55.3 \pm 4.9$ \\
Stack             & $62.3 \pm 5.3$   & $61.1 \pm 9.4$   & $56.2 \pm 5.0$ \\
Identity          & $54.7 \pm 0.2$   & $58.3 \pm 7.2$   & $69.7 \pm 13.1$ \\
Set               & $53.4 \pm 0.1$   & $59.6 \pm 6.4$   & $81.1 \pm 12.2$ \\
Union      & $48.6 \pm 0.7$ & $65.0 \pm 12.2$ & $59.8 \pm 9.0$ \\
Reverse    & $46.1 \pm 2.3$ & $57.8 \pm 7.0$ & $60.9 \pm 7.9$ \\
Delete     & $60.7 \pm 4.2$  & $59.2 \pm 8.1$ & $63.3 \pm 14.0$ \\

ECA               & $69.6 \pm 7.9$   & $63.1 \pm 14.4$  & $65.8 \pm 12.8$ \\
\bottomrule
\end{tabular}
\end{sc}
\end{small}
\end{center}
\vspace{1em}
\caption{
\textsc{Addition} task accuracy (mean $\pm$ standard deviation over 10 seeds) for models initialized with weights from different pretraining tasks. We report results for full model transfer, MLP-transfer, and attention-transfer.
}
\end{table}

% Reversed addition Component Transfers
\begin{table}[H]
\label{tab:reversedaddition-component-transfer}
\vskip 0.15in
\begin{center}
\begin{small}
\begin{sc}
\begin{tabular}{lccc}
\toprule
\textnormal{Pretraining task} & Full transfer & MLP only & Attention only \\\midrule
16-Dyck            & $83.5 \pm 0.6$   & $64.0 \pm 26.4$  & $49.1 \pm 20.3$ \\
8-Dyck shuffle        & $90.1 \pm 5.9$   & $65.8 \pm 24.8$  & $63.3 \pm 18.1$ \\
Stack             & $34.9 \pm 0.2$   & $74.4 \pm 24.7$  & $42.1 \pm 8.1$ \\
Identity          & $42.7 \pm 0.9$   & $71.7 \pm 29.2$  & $45.2 \pm 3.7$ \\
Set               & $44.6 \pm 5.1$   & $71.2 \pm 23.7$  & $54.4 \pm 10.4$ \\
Union      & $50.8 \pm 0.2$ & $72.3 \pm 29.6$ & $50.3 \pm 16.5$ \\
Reverse    & $46.8 \pm 1.3$ & $75.8 \pm 27.1$ & $44.6 \pm 3.4$ \\
Delete & $40.0 \pm 1.8$  & $55.2 \pm 23.0$ & $44.6 \pm 9.2$ \\
ECA               & $91.1 \pm 16.1$  & $70.5 \pm 31.6$  & $75.5 \pm 27.2$ \\
\bottomrule
\end{tabular}
\end{sc}
\end{small}
\end{center}
\vspace{1em}
\caption{
\textsc{Reversed addition} task accuracy (mean $\pm$ standard deviation over 10 seeds) for models initialized with weights from different pretraining tasks. We report results for full model transfer, MLP-transfer, and attention-transfer.
}
\end{table}

% Sorting Component Transfers
\begin{table}[H]
\label{tab:sorting-component-transfer}
\vskip 0.15in
\begin{center}
\begin{small}
\begin{sc}
\begin{tabular}{lccc}
\toprule
\textnormal{Pretraining task} & Full transfer & MLP only & Attention only \\\midrule
8-Dyck         & 98.7$\pm$0.3  & 72.8$\pm$3.1  & 71.4$\pm$5.7 \\
8-Dyck shuffle  & 84.1$\pm$5.7  & 78.2$\pm$8.6  & 62.9$\pm$6.7 \\
Stack          & 21.3$\pm$0.6  & 71.0$\pm$2.2  & 77.5$\pm$12.2 \\
Identity       & 19.9$\pm$0.5  & 74.5$\pm$8.1  & 91.3$\pm$10.1 \\
Set            & 93.5$\pm$1.6  & 73.5$\pm$1.5  & 98.1$\pm$2.8 \\
Union      & $16.9 \pm 0.5$ & $72.3 \pm 1.9$ &  $76.4 \pm 16.4$ \\
Reverse    & $16.7 \pm 0.5$ & $71.2 \pm 2.6$ & $82.1 \pm 15.1$ \\
Delete     & $20.1 \pm 0.6$ & $78.0 \pm 10.9$ & $81.3 \pm 24.3$ \\
ECA            & $76.9 \pm 1.4$  & 77.1$\pm$8.1 & 73.9$\pm$3.2 \\
\bottomrule
\end{tabular}
\end{sc}
\end{small}
\end{center}
\vspace{1em}
\caption{
\textsc{Sorting} task accuracy (mean $\pm$ standard deviation over 10 seeds) for models initialized with weights from different pretraining tasks. We report results for full model transfer, MLP-transfer, and attention-transfer.
}
\end{table}

\begin{table}[H]
\centering

% Weight Perturbation Experiments
\label{tab:perturbation-table}
\vspace{0.5em}
\begin{small}
\begin{tabular}{lccccc}
\toprule
\textnormal{Perturbation} & \textsc{Haystack} & \textsc{Addition} & \textsc{Reversed addition} & \textsc{Sorting} \\
\midrule
Pretrained   & $98.9 \pm 0.8$ & $87.8 \pm 4.2$ & $90.1 \pm 5.9$ & $98.7 \pm 0.3$ \\
Shuffled     & $17.2 \pm 12.7$ & $61.0 \pm 9.1$ & $82.9 \pm 23.5$ & $94.2 \pm 4.2$ \\
0.01 noise   & $98.6 \pm 1.7$ & $77.6 \pm 20.1$ & $74.0 \pm 21.0$ & $96.0 \pm 7.6$ \\
0.05 noise   & $50.8 \pm 30.5$ & $62.1 \pm 13.3$ & $91.0 \pm 15.7$ & $71.9 \pm 26.1$ \\
0.10 noise   & $32.9 \pm 6.1$ & $56.4 \pm 7.4$ & $83.6 \pm 21.5$ & $37.9 \pm 5.8$ \\
Random init  & $11.3 \pm 0.4$ & $59.1 \pm 7.0$ & $76.4 \pm 23.2$ & $82.7 \pm 11.6$ \\
\bottomrule
\end{tabular}
\vspace{1em}
\caption{
Mean accuracy ($\pm$ standard deviation over 10 seeds) across five algorithmic tasks under different perturbation conditions. Pretrained models were selected based on best individual performance per task: \textsc{Stack} (attention-transfer) for \textsc{Haystack}, \textsc{16-Dyck} for \textsc{Addition} (full-transfer), \textsc{8-Dyck shuffle} for \textsc{Reversed addition} (full-transfer), \textsc{8-Dyck} for \textsc{Sorting} (full-transfer).
}

\end{small}
\end{table}

\subsection{Semantic Data}
\label{app:more_results-semantic}
\begin{figure}[H]
    \centering
    \includegraphics[width=0.5\linewidth]{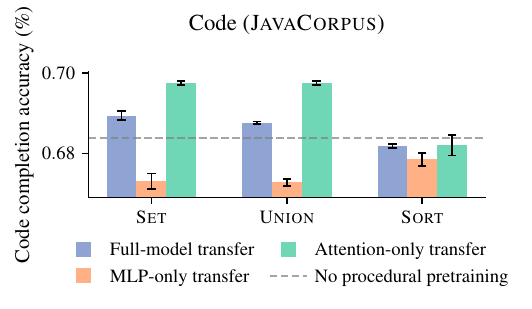}
    \caption{
    Token level code completion accuracy on \textsc{JavaCorpus} from \citep{lu2021codexglue}.
    We compare partial transfer of pretrained weights with full-model transfer. 
    This extends the partial transfer analysis from Figure~\ref{fig:sem_understanding_partial_transfer} in the main paper, showing Attention-only transfer is superior for code in isolation.
    }
    \label{fig:java_code_completion_acc}
\end{figure}

\begin{figure}[H]
    \centering
    \includegraphics[width=0.5\linewidth]{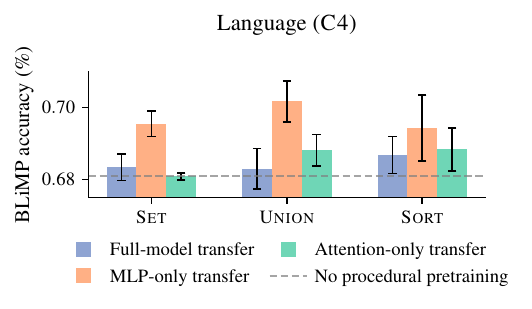}
    \caption{
    BLiMP accuracy~\citep{warstadt2020blimp} after training on \textsc{C4}. 
    We compare partial transfer of pretrained weights with full-model transfer. 
    Consistent with Figure~\ref{fig:sem_understanding_partial_transfer}, MLP-only transfer achieves the best performance on grammatical understanding.
    }
    \label{fig:c4_blimp_acc}
\end{figure}

\begin{figure}[H]
    \centering
    \includegraphics[width=1.0\linewidth]{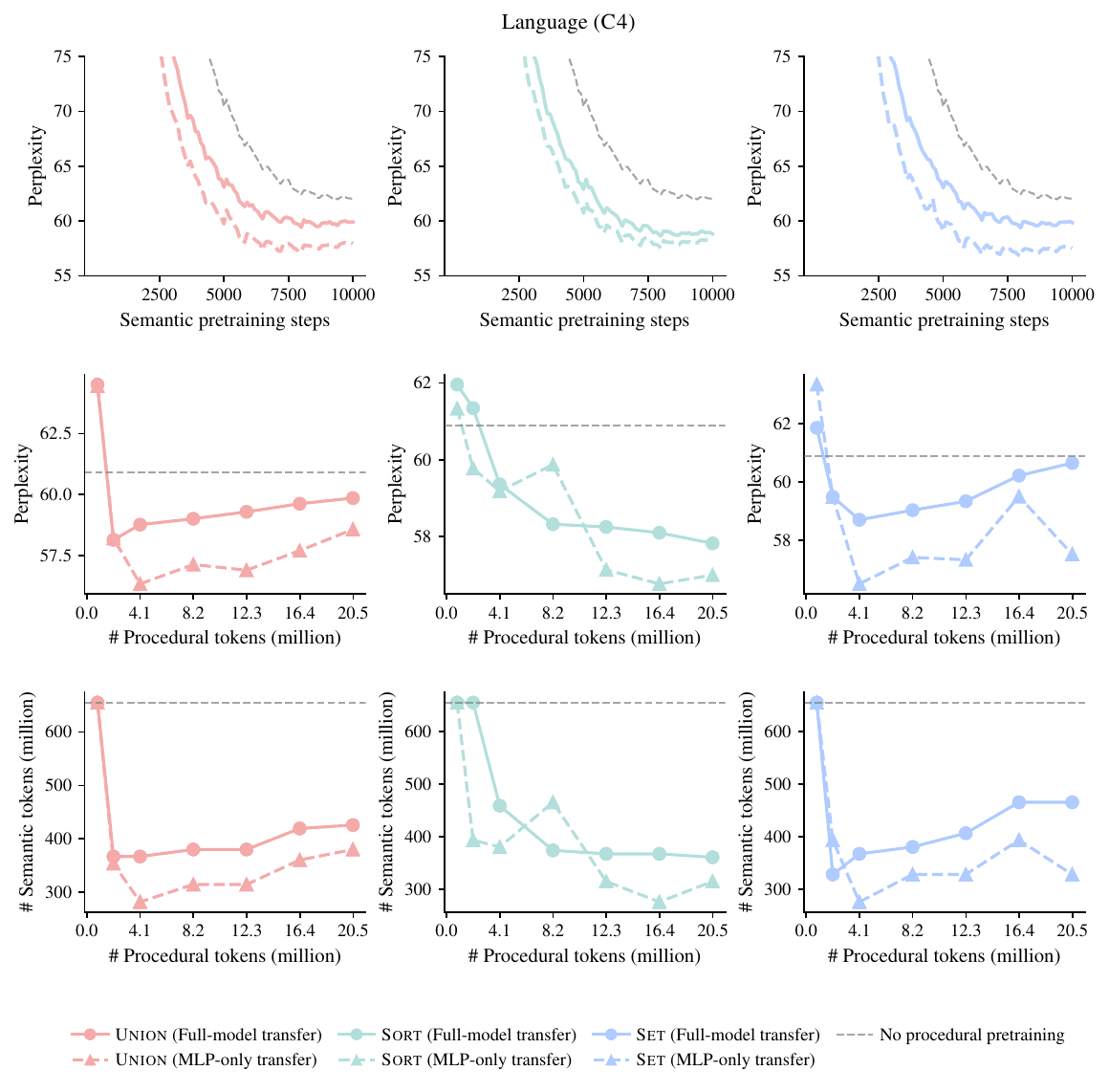}
    \caption{Comparison of MLP-only transfer and full-model transfer on \textsc{C4} for \textsc{Union}, \textsc{Sort} and \textsc{Set}. 
    (\textbf{Top}) Perplexity curves during semantic pretraining. 
    (\textbf{Middle})
    Additive setting results. 
    (\textbf{Bottom}) 
    Substitutive setting results. 
    Across all views, MLP-only transfer outperforms full transfer, confirming that procedurally pretrained MLP layers are especially effective for natural language.}
    \label{fig:c4_ffn_figure}
\end{figure}

\subsection{Weight Mixture}

\begin{table}[!h]
\scriptsize
    \vspace{-3pt}
    \centering
    \setlength{\tabcolsep}{3pt}
    \begin{tabular}{lcccc}
    \toprule
    %\textnormal{Pretraining configuration}
    & \textsc{Haystack} & \textsc{Addition} & \textsc{Reversed addition} & \textsc{Sort} \\
    \midrule
    \textnormal{No procedural pretraining}       & $11.3_{\pm 0.4}$  & $59.1_{\pm 7.0}$   & $76.4_{\pm 23.2}$  & $82.7_{\pm 11.6}$ \\

    \midrule
    \textsc{Set} ~~\textnormal{(full-model transfer)}       & $18.9_{\pm 26.6}$  & $53.4_{\pm 0.1}$   & $44.6_{\pm 5.1}$  & $93.5_{\pm 1.6}$ \\
    \textsc{Set} ~~\textnormal{(attention-only transfer)}      & $88.9_{\pm 27.1}$  & $\mathbf{81.1}_{\pm 12.2}$   & $54.4_{\pm 10.4}$  & $98.1_{\pm 2.8}$ \\
    \midrule
    \textsc{ECA} ~~\textnormal{(full-model transfer)}       & $10.5_{\pm 0.5}$   & $69.6_{\pm 7.9}$   & $\mathbf{91.0}_{\pm 16.1}$  & $76.9_{\pm 1.4}$ \\
    \textsc{ECA} ~~\textnormal{(MLP-only transfer)}           & $8.71_{\pm 1.0}$   & $63.1_{\pm 14.4}$  & $70.5_{\pm 31.6}$  & $77.1_{\pm 8.1}$ \\
    \midrule
    \textsc{Set} \textnormal{(attention)} \;+\; ECA \textnormal{(MLP)}~~~~~~~~ & $\mathbf{94.4}_{\pm 2.5}$  & $\underline{80.3}_{\pm 13.9}$   & $\underline{82.9}_{\pm 16.9}$  & $\mathbf{99.4}_{\pm 0.2}$ \\
    \bottomrule
    \end{tabular}
    \vspace{1em}
    \captionof{table}{\textbf{Pretrained models combined at the weight level.}
We combine \textsc{Set}-pretrained attention layers with \textsc{ECA}-pretrained MLPs (last row).
This yields the strong performance across all four tasks, whereas single-source models show weaknesses in at least one task.}
    \label{tab:mixture_weights_full}
    \vspace{-5pt}
\end{table}

%% file: 0_main.bbl
\begin{thebibliography}{66}
\providecommand{\natexlab}[1]{#1}
\providecommand{\url}[1]{\texttt{#1}}
\expandafter\ifx\csname urlstyle\endcsname\relax
  \providecommand{\doi}[1]{doi: #1}\else
  \providecommand{\doi}{doi: \begingroup \urlstyle{rm}\Url}\fi

\bibitem[Abnar et~al.(2020)Abnar, Dehghani, and Zuidema]{abnar2020transferring}
Abnar, S., Dehghani, M., and Zuidema, W.
\newblock Transferring inductive biases through knowledge distillation.
\newblock \emph{arXiv preprint arXiv:2006.00555}, 2020.

\bibitem[Allamanis \& Sutton(2013)Allamanis and Sutton]{allamanis2013mining}
Allamanis, M. and Sutton, C.
\newblock Mining source code repositories at massive scale using language modeling.
\newblock In \emph{2013 10th working conference on mining software repositories (MSR)}. IEEE, 2013.

\bibitem[Aryabumi et~al.(2024)Aryabumi, Su, Ma, Morisot, Zhang, Locatelli, Fadaee, {\"U}st{\"u}n, and Hooker]{aryabumi2024code}
Aryabumi, V., Su, Y., Ma, R., Morisot, A., Zhang, I., Locatelli, A., Fadaee, M., {\"U}st{\"u}n, A., and Hooker, S.
\newblock To code, or not to code? exploring impact of code in pre-training.
\newblock \emph{arXiv preprint arXiv:2408.10914}, 2024.

\bibitem[Austin et~al.(2021)Austin, Odena, Nye, Bosma, Michalewski, Dohan, Jiang, Cai, Terry, Le, and Sutton]{austin2021programsynthesislargelanguage}
Austin, J., Odena, A., Nye, M., Bosma, M., Michalewski, H., Dohan, D., Jiang, E., Cai, C., Terry, M., Le, Q., and Sutton, C.
\newblock Program synthesis with large language models, 2021.
\newblock URL \url{https://arxiv.org/abs/2108.07732}.

\bibitem[Balestriero \& Huang(2024)Balestriero and Huang]{balestriero2024perception}
Balestriero, R. and Huang, H.
\newblock For perception tasks: The cost of llm pretraining by next-token prediction outweigh its benefits.
\newblock In \emph{NeurIPS Workshop: Self-Supervised Learning-Theory and Practice}, 2024.

\bibitem[Baradad et~al.(2021)Baradad, Wulff, Wang, Isola, and Torralba]{baradad2021learning}
Baradad, M., Wulff, J., Wang, T., Isola, P., and Torralba, A.
\newblock Learning to see by looking at noise.
\newblock \emph{arXiv preprint arXiv:2106.05963}, 2021.

\bibitem[Baradad et~al.(2022)Baradad, Chen, Wulff, Wang, Feris, Torralba, and Isola]{baradad2022procedural}
Baradad, M., Chen, C.-F., Wulff, J., Wang, T., Feris, R., Torralba, A., and Isola, P.
\newblock Procedural image programs for representation learning.
\newblock \emph{arXiv preprint arXiv:2211.16412}, 2022.

\bibitem[Biderman et~al.(2023)Biderman, Schoelkopf, Anthony, Bradley, O’Brien, Hallahan, Khan, Purohit, Prashanth, Raff, et~al.]{biderman2023pythia}
Biderman, S., Schoelkopf, H., Anthony, Q.~G., Bradley, H., O’Brien, K., Hallahan, E., Khan, M.~A., Purohit, S., Prashanth, U.~S., Raff, E., et~al.
\newblock Pythia: A suite for analyzing large language models across training and scaling.
\newblock In \emph{International Conference on Machine Learning}, pp.\  2397--2430. PMLR, 2023.

\bibitem[Bloem(2025)]{bloem2025universalpretrainingiteratedrandom}
Bloem, P.
\newblock Universal pre-training by iterated random computation, 2025.
\newblock URL \url{https://arxiv.org/abs/2506.20057}.

\bibitem[Chan et~al.(2022)Chan, Santoro, Lampinen, Wang, Singh, Richemond, McClelland, and Hill]{chan2022data}
Chan, S., Santoro, A., Lampinen, A., Wang, J., Singh, A., Richemond, P., McClelland, J., and Hill, F.
\newblock Data distributional properties drive emergent in-context learning in transformers.
\newblock \emph{Advances in Neural Information Processing Systems}, 2022.

\bibitem[Charton \& Kempe(2024)Charton and Kempe]{charton2024emergent}
Charton, F. and Kempe, J.
\newblock Emergent properties with repeated examples.
\newblock \emph{arXiv preprint arXiv:2410.07041}, 2024.

\bibitem[Chiang \& Lee(2022)Chiang and Lee]{chiang2022transferability}
Chiang, C.-H. and Lee, H.-y.
\newblock On the transferability of pre-trained language models: A study from artificial datasets.
\newblock In \emph{Proceedings of the AAAI Conference on Artificial Intelligence}, 2022.

\bibitem[Clark et~al.(2018)Clark, Cowhey, Etzioni, Khot, Sabharwal, Schoenick, and Tafjord]{clark2018think}
Clark, P., Cowhey, I., Etzioni, O., Khot, T., Sabharwal, A., Schoenick, C., and Tafjord, O.
\newblock Think you have solved question answering? try arc, the ai2 reasoning challenge.
\newblock \emph{arXiv preprint arXiv:1803.05457}, 2018.

\bibitem[Conmy et~al.(2023)Conmy, Mavor-Parker, Lynch, Heimersheim, and Garriga-Alonso]{conmy2023automated}
Conmy, A., Mavor-Parker, A.~N., Lynch, A., Heimersheim, S., and Garriga-Alonso, A.
\newblock Towards automated circuit discovery for mechanistic interpretability.
\newblock In \emph{Thirty-seventh Conference on Neural Information Processing Systems}, 2023.

\bibitem[Dong et~al.(2025)Dong, Noci, Khodak, and Li]{dong2025attention}
Dong, Y., Noci, L., Khodak, M., and Li, M.
\newblock Is random attention sufficient for sequence modeling? disentangling trainable components in the transformer, 2025.
\newblock URL \url{https://arxiv.org/abs/2506.01115}.

\bibitem[Fan et~al.(2023)Fan, Pagliardini, and Jaggi]{fan2023doge}
Fan, S., Pagliardini, M., and Jaggi, M.
\newblock Doge: Domain reweighting with generalization estimation.
\newblock \emph{arXiv preprint arXiv:2310.15393}, 2023.

\bibitem[Gao et~al.(2020)Gao, Biderman, Black, Golding, Hoppe, Foster, Phang, He, Thite, Nabeshima, et~al.]{gao2020pile}
Gao, L., Biderman, S., Black, S., Golding, L., Hoppe, T., Foster, C., Phang, J., He, H., Thite, A., Nabeshima, N., et~al.
\newblock The {Pile}: An {800GB} dataset of diverse text for language modeling.
\newblock \emph{arXiv preprint arXiv:2101.00027}, 2020.

\bibitem[Geva et~al.(2020)Geva, Schuster, Berant, and Levy]{geva2020transformer}
Geva, M., Schuster, R., Berant, J., and Levy, O.
\newblock Transformer feed-forward layers are key-value memories.
\newblock \emph{arXiv preprint arXiv:2012.14913}, 2020.

\bibitem[Goodale et~al.(2025)Goodale, Mascarenhas, and Lakretz]{goodale2025meta}
Goodale, M., Mascarenhas, S., and Lakretz, Y.
\newblock Meta-learning neural mechanisms rather than bayesian priors.
\newblock \emph{arXiv preprint arXiv:2503.16048}, 2025.

\bibitem[Grau-Moya et~al.(2024)Grau-Moya, Genewein, Hutter, Orseau, Del{\'e}tang, Catt, Ruoss, Wenliang, Mattern, Aitchison, et~al.]{grau2024learning}
Grau-Moya, J., Genewein, T., Hutter, M., Orseau, L., Del{\'e}tang, G., Catt, E., Ruoss, A., Wenliang, L.~K., Mattern, C., Aitchison, M., et~al.
\newblock Learning universal predictors.
\newblock \emph{arXiv preprint arXiv:2401.14953}, 2024.

\bibitem[Han et~al.(2025)Han, Pari, Gershman, and Agrawal]{han2025position}
Han, S., Pari, J., Gershman, S.~J., and Agrawal, P.
\newblock Position: General intelligence requires reward-based pretraining.
\newblock In \emph{{Proceedings of the International Conference on Machine Learning} Position Paper Track}, 2025.

\bibitem[He et~al.(2023)He, Blackwood, Panda, McAuley, and Feris]{he2023synthetic}
He, Z., Blackwood, G., Panda, R., McAuley, J., and Feris, R.
\newblock Synthetic pre-training tasks for neural machine translation.
\newblock In \emph{Findings of the Association for Computational Linguistics}, 2023.

\bibitem[Hu et~al.(2025)Hu, Petty, Shi, Merrill, and Linzen]{hu2025between}
Hu, M.~Y., Petty, J., Shi, C., Merrill, W., and Linzen, T.
\newblock Between circuits and chomsky: Pre-pretraining on formal languages imparts linguistic biases.
\newblock In \emph{Proceedings of the 63rd Annual Meeting of the Association for Computational Linguistics (Long Papers)}, 2025.

\bibitem[Huang et~al.(2020)Huang, Perez, Ba, and Volkovs]{huang2020improving}
Huang, X.~S., Perez, F., Ba, J., and Volkovs, M.
\newblock Improving transformer optimization through better initialization.
\newblock In \emph{Proceedings of the International Conference on Machine Learning}, 2020.

\bibitem[HuggingFace(2022)]{codeparrot}
HuggingFace.
\newblock Codeparrot dataset cleaned, 2022.

\bibitem[Huh et~al.(2024)Huh, Cheung, Wang, and Isola]{huh2024platonic}
Huh, M., Cheung, B., Wang, T., and Isola, P.
\newblock The platonic representation hypothesis.
\newblock \emph{arXiv preprint arXiv:2405.07987}, 2024.

\bibitem[Jesus et~al.(2021)Jesus, Antunes, da~Costa, Dorogovtsev, Mendes, and Aguiar]{jesus2021effect}
Jesus, R.~J., Antunes, M.~L., da~Costa, R.~A., Dorogovtsev, S.~N., Mendes, J.~F., and Aguiar, R.~L.
\newblock Effect of initial configuration of weights on training and function of artificial neural networks.
\newblock \emph{Mathematics}, 9\penalty0 (18):\penalty0 2246, 2021.

\bibitem[Kaplan et~al.(2020)Kaplan, McCandlish, Henighan, Brown, Chess, Child, Gray, Radford, Wu, and Amodei]{kaplan2020scaling}
Kaplan, J., McCandlish, S., Henighan, T., Brown, T.~B., Chess, B., Child, R., Gray, S., Radford, A., Wu, J., and Amodei, D.
\newblock Scaling laws for neural language models.
\newblock \emph{arXiv preprint arXiv:2001.08361}, 2020.

\bibitem[Krishna et~al.(2023)Krishna, Garg, Bigham, and Lipton]{krishna2023downstream}
Krishna, K., Garg, S., Bigham, J., and Lipton, Z.
\newblock Downstream datasets make surprisingly good pretraining corpora.
\newblock In \emph{Proceedings of the 61st Annual Meeting of the Association for Computational Linguistics}, 2023.

\bibitem[Kumar et~al.(2025)Kumar, Clune, Lehman, and Stanley]{kumar2025questioning}
Kumar, A., Clune, J., Lehman, J., and Stanley, K.~O.
\newblock Questioning representational optimism in deep learning: The fractured entangled representation hypothesis.
\newblock \emph{arXiv preprint arXiv:2505.11581}, 2025.

\bibitem[Lindemann et~al.(2024)Lindemann, Koller, and Titov]{lindemann2024sip}
Lindemann, M., Koller, A., and Titov, I.
\newblock Sip: Injecting a structural inductive bias into a seq2seq model by simulation.
\newblock In \emph{Proceedings of the 62nd Annual Meeting of the Association for Computational Linguistics}, 2024.

\bibitem[Liu et~al.(2023)Liu, Ash, Goel, Krishnamurthy, and Zhang]{liu2023exposing}
Liu, B., Ash, J., Goel, S., Krishnamurthy, A., and Zhang, C.
\newblock Exposing attention glitches with flip-flop language modeling.
\newblock \emph{Advances in Neural Information Processing Systems}, 36:\penalty0 25549--25583, 2023.

\bibitem[Longpre et~al.(2024)Longpre, Yauney, Reif, Lee, Roberts, Zoph, Zhou, Wei, Robinson, Mimno, et~al.]{longpre2024pretrainer}
Longpre, S., Yauney, G., Reif, E., Lee, K., Roberts, A., Zoph, B., Zhou, D., Wei, J., Robinson, K., Mimno, D., et~al.
\newblock A pretrainer’s guide to training data: Measuring the effects of data age, domain coverage, quality, \& toxicity.
\newblock In \emph{Proceedings of the Conference of the North American Chapter of the Association for Computational Linguistics}, 2024.

\bibitem[Lu et~al.(2021)Lu, Guo, Ren, Huang, Svyatkovskiy, Blanco, Clement, Drain, Jiang, Tang, Li, Zhou, Shou, Zhou, Tufano, Gong, Zhou, Duan, Sundaresan, Deng, Fu, and Liu]{lu2021codexglue}
Lu, S., Guo, D., Ren, S., Huang, J., Svyatkovskiy, A., Blanco, A., Clement, C., Drain, D., Jiang, D., Tang, D., Li, G., Zhou, L., Shou, L., Zhou, L., Tufano, M., Gong, M., Zhou, M., Duan, N., Sundaresan, N., Deng, S.~K., Fu, S., and Liu, S.
\newblock Codexglue: A machine learning benchmark dataset for code understanding and generation.
\newblock \emph{arXiv preprint arXiv:2102.04664}, 2021.

\bibitem[McCoy \& Griffiths(2023)McCoy and Griffiths]{mccoy2023modeling}
McCoy, R.~T. and Griffiths, T.~L.
\newblock Modeling rapid language learning by distilling bayesian priors into artificial neural networks.
\newblock \emph{arXiv preprint arXiv:2305.14701}, 2023.

\bibitem[Merity et~al.(2016)Merity, Xiong, Bradbury, and Socher]{merity2016pointer}
Merity, S., Xiong, C., Bradbury, J., and Socher, R.
\newblock Pointer sentinel mixture models.
\newblock \emph{arXiv preprint arXiv:1609.07843}, 2016.

\bibitem[M{\"u}ller et~al.(2021)M{\"u}ller, Hollmann, Arango, Grabocka, and Hutter]{muller2021transformers}
M{\"u}ller, S., Hollmann, N., Arango, S.~P., Grabocka, J., and Hutter, F.
\newblock Transformers can do bayesian inference.
\newblock \emph{arXiv preprint arXiv:2112.10510}, 2021.

\bibitem[Nakamura et~al.(2024)Nakamura, Tadokoro, Yamada, Asano, Laina, Rupprecht, Inoue, Yokota, and Kataoka]{nakamura2024scaling}
Nakamura, R., Tadokoro, R., Yamada, R., Asano, Y.~M., Laina, I., Rupprecht, C., Inoue, N., Yokota, R., and Kataoka, H.
\newblock Scaling backwards: Minimal synthetic pre-training?
\newblock \emph{arXiv preprint arXiv:2408.00677}, 2024.

\bibitem[Nikankin et~al.(2025)Nikankin, Reusch, Mueller, and Belinkov]{nikankin2025arithmetic}
Nikankin, Y., Reusch, A., Mueller, A., and Belinkov, Y.
\newblock Arithmetic without algorithms: Language models solve math with a bag of heuristics.
\newblock In \emph{The Thirteenth International Conference on Learning Representations}, 2025.

\bibitem[Papadimitriou \& Jurafsky(2023)Papadimitriou and Jurafsky]{papadimitriou2023injecting}
Papadimitriou, I. and Jurafsky, D.
\newblock Injecting structural hints: Using language models to study inductive biases in language learning.
\newblock \emph{arXiv preprint arXiv:2304.13060}, 2023.

\bibitem[Petty et~al.(2024)Petty, van Steenkiste, and Linzen]{petty2024does}
Petty, J., van Steenkiste, S., and Linzen, T.
\newblock How does code pretraining affect language model task performance?
\newblock \emph{arXiv preprint arXiv:2409.04556}, 2024.

\bibitem[Pouransari et~al.(2025)Pouransari, Grangier, Thomas, Kirchhof, and Tuzel]{pouransari2025pretraining}
Pouransari, H., Grangier, D., Thomas, C., Kirchhof, M., and Tuzel, O.
\newblock Pretraining with hierarchical memories: separating long-tail and common knowledge.
\newblock \emph{arXiv preprint arXiv:2510.02375}, 2025.

\bibitem[Radford et~al.(2019)Radford, Wu, Child, Luan, Amodei, Sutskever, et~al.]{radford2019language}
Radford, A., Wu, J., Child, R., Luan, D., Amodei, D., Sutskever, I., et~al.
\newblock Language models are unsupervised multitask learners.
\newblock \emph{OpenAI blog}, 2019.

\bibitem[Raffel et~al.(2020)Raffel, Shazeer, Roberts, Lee, Narang, Matena, Zhou, Li, and Liu]{raffel2020exploring}
Raffel, C., Shazeer, N., Roberts, A., Lee, K., Narang, S., Matena, M., Zhou, Y., Li, W., and Liu, P.~J.
\newblock Exploring the limits of transfer learning with a unified text-to-text transformer.
\newblock \emph{Journal of machine learning research}, 21\penalty0 (140):\penalty0 1--67, 2020.

\bibitem[Raychev et~al.(2016)Raychev, Bielik, and Vechev]{raychev2016probabilistic}
Raychev, V., Bielik, P., and Vechev, M.
\newblock Probabilistic model for code with decision trees.
\newblock \emph{ACM SIGPLAN Notices}, 51\penalty0 (10):\penalty0 731--747, 2016.

\bibitem[Ri \& Tsuruoka(2022)Ri and Tsuruoka]{ri2022pretraining}
Ri, R. and Tsuruoka, Y.
\newblock Pretraining with artificial language: Studying transferable knowledge in language models.
\newblock \emph{arXiv preprint arXiv:2203.10326}, 2022.

\bibitem[Ruis et~al.(2024)Ruis, Mozes, Bae, Kamalakara, Talupuru, Locatelli, Kirk, Rockt{\"a}schel, Grefenstette, and Bartolo]{ruis2024procedural}
Ruis, L., Mozes, M., Bae, J., Kamalakara, S.~R., Talupuru, D., Locatelli, A., Kirk, R., Rockt{\"a}schel, T., Grefenstette, E., and Bartolo, M.
\newblock Procedural knowledge in pretraining drives reasoning in large language models.
\newblock \emph{arXiv preprint arXiv:2411.12580}, 2024.

\bibitem[Saxton et~al.(2019)Saxton, Grefenstette, Hill, and Kohli]{saxton2018analysing}
Saxton, D., Grefenstette, E., Hill, F., and Kohli, P.
\newblock Analysing mathematical reasoning abilities of neural models.
\newblock In \emph{International Conference on Learning Representations}, 2019.

\bibitem[Shinnick et~al.(2026)Shinnick, Jiang, Saratchandran, Teney, and Hengel]{shinnick2025can}
Shinnick, Z., Jiang, L., Saratchandran, H., Teney, D., and Hengel, A. v.~d.
\newblock Can you learn to see without images? procedural warm-up for vision transformers.
\newblock In \emph{Proceedings of the IEEE/CVF Conference on Computer Vision and Pattern Recognition}, 2026.

\bibitem[Smith \& Gasser(2005)Smith and Gasser]{smith2005development}
Smith, L. and Gasser, M.
\newblock The development of embodied cognition: Six lessons from babies.
\newblock \emph{Artificial life}, 11\penalty0 (1-2):\penalty0 13--29, 2005.

\bibitem[Teney et~al.(2024)Teney, Nicolicioiu, Hartmann, and Abbasnejad]{teney2024neural}
Teney, D., Nicolicioiu, A.~M., Hartmann, V., and Abbasnejad, E.
\newblock Neural redshift: Random networks are not random functions.
\newblock In \emph{Proceedings of the IEEE/CVF Conference on Computer Vision and Pattern Recognition}, 2024.

\bibitem[Teney et~al.(2025)Teney, Jiang, Gogianu, and Abbasnejad]{teney2025we}
Teney, D., Jiang, L., Gogianu, F., and Abbasnejad, E.
\newblock Do we always need the simplicity bias? looking for optimal inductive biases in the wild.
\newblock In \emph{Proceedings of the IEEE/CVF Conference on Computer Vision and Pattern Recognition}, 2025.

\bibitem[Trockman \& Kolter(2023)Trockman and Kolter]{trockman2023mimetic}
Trockman, A. and Kolter, J.~Z.
\newblock Mimetic initialization of self-attention layers.
\newblock \emph{arXiv preprint arXiv:2305.09828}, 2023.

\bibitem[Wang et~al.(2019)Wang, Singh, Michael, Hill, Levy, and Bowman]{wang2018glue}
Wang, A., Singh, A., Michael, J., Hill, F., Levy, O., and Bowman, S.~R.
\newblock {GLUE}: A multi-task benchmark and analysis platform for natural language understanding.
\newblock In \emph{International Conference on Learning Representations}, 2019.
\newblock URL \url{https://openreview.net/forum?id=rJ4km2R5t7}.

\bibitem[Wang et~al.(2022)Wang, Ko, and Agrawal]{wang2022visual}
Wang, Y., Ko, C.-Y., and Agrawal, P.
\newblock Visual pre-training for navigation: What can we learn from noise?
\newblock \emph{arXiv preprint arXiv:2207.00052}, 2022.

\bibitem[Wang et~al.(2023)Wang, Wang, Dong, and Ross]{wang2023pretraining}
Wang, Z., Wang, C., Dong, Z., and Ross, K.
\newblock Pre-training with synthetic data helps offline reinforcement learning.
\newblock \emph{arXiv preprint arXiv:2310.00771}, 2023.

\bibitem[Warstadt et~al.(2020)Warstadt, Parrish, Liu, Mohananey, Peng, Wang, and Bowman]{warstadt2020blimp}
Warstadt, A., Parrish, A., Liu, H., Mohananey, A., Peng, W., Wang, S.-F., and Bowman, S.~R.
\newblock Blimp: The benchmark of linguistic minimal pairs for english.
\newblock \emph{Transactions of the Association for Computational Linguistics}, 2020.

\bibitem[Wu et~al.(2021)Wu, Rabe, Li, Ba, Grosse, and Szegedy]{wu2021lime}
Wu, Y., Rabe, M.~N., Li, W., Ba, J., Grosse, R.~B., and Szegedy, C.
\newblock Lime: Learning inductive bias for primitives of mathematical reasoning.
\newblock In \emph{Proceedings of the International Conference on Machine Learning}, 2021.

\bibitem[Wu et~al.(2022)Wu, Li, and Liang]{wu2022insights}
Wu, Y., Li, F., and Liang, P.~S.
\newblock Insights into pre-training via simpler synthetic tasks.
\newblock \emph{Advances in Neural Information Processing Systems}, 2022.

\bibitem[Xie et~al.(2023)Xie, Pham, Dong, Du, Liu, Lu, Liang, Le, Ma, and Yu]{xie2023doremi}
Xie, S.~M., Pham, H., Dong, X., Du, N., Liu, H., Lu, Y., Liang, P.~S., Le, Q.~V., Ma, T., and Yu, A.~W.
\newblock Doremi: Optimizing data mixtures speeds up language model pretraining.
\newblock \emph{Advances in Neural Information Processing Systems}, 2023.

\bibitem[Xie et~al.(2025)Xie, Tonin, and Cevher]{xie2025chameleon}
Xie, W., Tonin, F., and Cevher, V.
\newblock Chameleon: A flexible data-mixing framework for language model pretraining and finetuning.
\newblock In \emph{Proceedings of the International Conference on Machine Learning}, 2025.

\bibitem[Xu \& Chen(2025)Xu and Chen]{xu2025filtering}
Xu, R. and Chen, K.
\newblock Filtering with self-attention and storing with mlp: One-layer transformers can provably acquire and extract knowledge.
\newblock \emph{arXiv preprint arXiv:2508.00901}, 2025.

\bibitem[Xu et~al.(2023)Xu, Chen, Vishniakov, Yin, Shen, Darrell, Liu, and Liu]{xu2023initializing}
Xu, Z., Chen, Y., Vishniakov, K., Yin, Y., Shen, Z., Darrell, T., Liu, L., and Liu, Z.
\newblock Initializing models with larger ones.
\newblock \emph{arXiv preprint arXiv:2311.18823}, 2023.

\bibitem[Zellers et~al.(2019)Zellers, Holtzman, Bisk, Farhadi, and Choi]{zellers2019hellaswag}
Zellers, R., Holtzman, A., Bisk, Y., Farhadi, A., and Choi, Y.
\newblock Hellaswag: Can a machine really finish your sentence?
\newblock In \emph{Proceedings of the 57th annual meeting of the association for computational linguistics}, pp.\  4791--4800, 2019.

\bibitem[Zhang et~al.(2023)Zhang, Lepori, and Pavlick]{zhang2023instilling}
Zhang, E., Lepori, M.~A., and Pavlick, E.
\newblock Instilling inductive biases with subnetworks.
\newblock \emph{arXiv preprint arXiv:2310.10899}, 2023.

\bibitem[Zhang et~al.(2024)Zhang, Patel, Rizvi, Liu, He, Karbasi, Zappala, and van Dijk]{zhang2024intelligence}
Zhang, S., Patel, A., Rizvi, S.~A., Liu, N., He, S., Karbasi, A., Zappala, E., and van Dijk, D.
\newblock Intelligence at the edge of chaos.
\newblock \emph{arXiv preprint arXiv:2410.02536}, 2024.

\end{thebibliography}
